\pdfoutput=1
\documentclass[10pt,twocolumn,letterpaper]{article}

\usepackage[pagenumbers]{cvpr}

\usepackage{multirow}
\usepackage{makecell}
\usepackage{amsthm}
\usepackage{tikz}
\usetikzlibrary{arrows.meta,positioning,fit,backgrounds,calc}
\theoremstyle{plain}
\newtheorem{proposition}{Proposition}

\newtheorem{lemma}{Lemma}
\theoremstyle{definition}

\newcommand{\cross}{\kappa}
\newcommand{\method}{FloorSight\xspace}
\newcommand{\prog}{\mathcal{P}}
\newcommand{\G}{\mathcal{G}}
\newcommand{\V}{\mathcal{V}}
\newcommand{\spec}{\mathcal{S}}

\definecolor{linkblue}{rgb}{0.21,0.49,0.74}
\usepackage[breaklinks,colorlinks,allcolors=linkblue]{hyperref}
\hypersetup{
    pdftitle={FloorSight: Constraint-Guided Generation of Floorplan Room Graphs},
    pdfauthor={Harmanjot Singh, Konstantinos Divriotis, Michalis Vazirgiannis}
}
\title{FloorSight: Constraint-Guided Generation of Floorplan Room Graphs}

\author{%
Harmanjot Singh$^{1,}$\thanks{Corresponding author.} \qquad
Konstantinos Divriotis$^{1}$ \qquad
Michalis Vazirgiannis$^{1,2}$\\[0.4em]
{\normalsize $^{1}$Mohamed bin Zayed University of Artificial Intelligence}\\
{\normalsize $^{2}$LIX, CNRS, \'Ecole Polytechnique, IP Paris}\\[0.3em]
{\tt\small \{harmanjot.singh, konstantinos.d, michalis.vazirgiannis\}@mbzuai.ac.ae}
}

\begin{document}
\maketitle
\begin{abstract}
Floorplan design begins with topology. A room graph records which rooms share a wall and which connect by a door or passage, and it fixes circulation, privacy and zoning before any geometry is drawn. Most learned layout generators take this graph as input. The requirements a user places on a home, such as an en-suite bathroom or a kitchen reachable without crossing a bedroom, are properties of this graph that can be checked exactly, so graph generation is the stage at which they can be guaranteed. We study how to generate room graphs that meet a user's brief, especially briefs that are rare in the training data, where published graph generators reach a success rate of at most 0.27. A generator that builds a graph step by step does not know which requirements are already met, which are still open and which it has made impossible. We show that for floorplan requirements this status can be computed exactly at every step by checking the sparsest and densest completions of the partial graph, and that standard message-passing encoders provably cannot infer it in general. We therefore feed the status to a masked graph generator and train it on real plans alone, with no reward or solver. Without a certifier, this lifts success on rare briefs from 0.10 to 0.47, while a more expressive encoder without the status stays at 0.11. With exact certifiers that guarantee the brief, the status cuts repair edits by 45\% and the draws needed by filtering about fivefold. Downstream, certified graphs nearly double the briefs met in floorplans drawn by a diffusion-based layout generator, compared with unconstrained sampling. Supplying the exact feasibility status thus lets a learned generator meet rare user requirements under a guarantee, with far less correction.
\end{abstract}
\section{Introduction}
\label{sec:intro}

\paragraph{Floorplans are planned as graphs, and users express requirements.}
A residential floorplan starts as a bubble diagram, a graph of which rooms share a wall and which open onto each other. This graph fixes circulation, privacy and zoning, and most learned layout generators take it as input~\cite{nauata2021houseganpp,hu2020graph2plan,shabani2023housediffusion}. Requirements such as an en-suite bathroom or a kitchen reachable without crossing a bedroom form a \emph{brief}. A brief can be checked exactly on this graph, so it can be guaranteed when the graph is generated.

\paragraph{Generators fail on rare briefs.}
We build a room-graph corpus from ResPlan~\cite{abouagour2025resplan} and MagicPlan~\cite{hosseini2023puzzlefusion} (\cref{sec:data}) and test on the 584 hard MagicPlan programs (room lists), whose brief is rare, as fewer than 5\% of the training plans with the same rooms meet it. Five published graph generators conditioned on the brief~\cite{you2018graphrnn,li2018learning,xia2024interactive,vignac2023digress,madeira2024construct} reach a success rate (valid and meeting the brief) of at most 0.27 and our own masked generator 0.105, although 0.834 of these briefs can be met (\cref{tab:published,tab:certifiers}). Constrained decoding and projection act on the sampler~\cite{madeira2024construct,sharma2024prodigy,zhang2026nspsg,jiang2023layoutformerpp,inoue2023layoutdm}, but never tell the network which requirements it can still meet.

\begin{figure*}[t]
  \centering
  \includegraphics[width=0.97\textwidth]{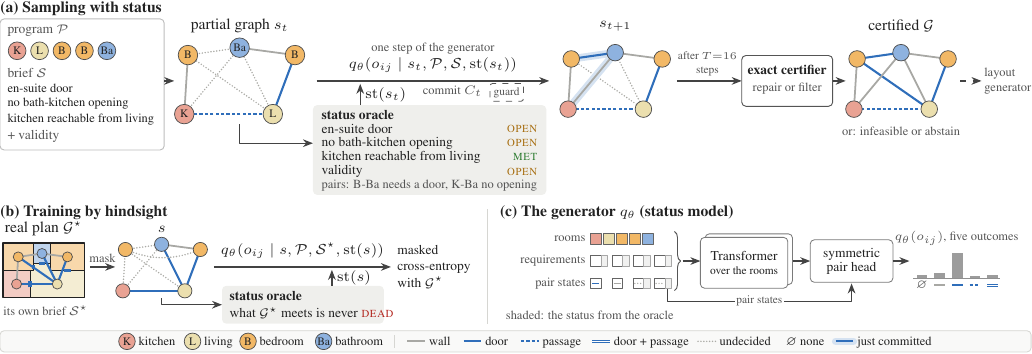}
  \caption{\textbf{\method.} (a) Sampling. Given a program $\prog$ and a brief $\spec$, the generator $q_\theta$ decides random sets $C_t$ of undecided pairs over $T{=}16$ steps. After each step the status oracle states whether each requirement is \textsc{met}, \textsc{open} or \textsc{dead} and which outcomes each undecided pair may still take. An exact certifier returns a certified graph, reports the brief infeasible or abstains. (b) Hindsight training on real plans with their own brief $\spec^\star$, which is never \textsc{dead}. (c) The status enters the generator as requirement flags and allowed outcomes (shaded).}
  \label{fig:method}
\end{figure*}

\paragraph{Telling the generator its feasibility status.}
For a partial graph, each requirement is already met, still open, or dead, meaning that no completion can meet it. Most floorplan requirements are monotone, as added walls and openings either never break them (connectivity, reachability) or never help them (bans, caps). For these the status follows exactly from the sparsest and densest completions (\cref{lem:status}), yet encoders bounded by 1-WL~\cite{xu2019powerful} cannot infer it in general (\cref{sec:method-status}). We therefore give the status to the generator as an input (\cref{fig:method}) and train it by hindsight on real plans, using the requirements each plan itself meets. No reward or solver is needed, since a plan's own outcomes never kill a requirement it meets (\cref{lem:hindsight}).

\paragraph{Exact certifiers do less work.}
Sampling alone guarantees nothing, so we pair the model with an exact certifier, either a sequential status guard that vetoes outcomes that would kill a requirement, minimum-edit SMT repair as in NSPSG~\cite{zhang2026nspsg}, or generate-and-filter. The status input reduces the work of all three. The guard's departure from exact conditional sampling falls more than threefold, repair needs 45\% fewer edits, and filtering needs about five times fewer draws on hard briefs (\cref{sec:exp-main}).

\paragraph{Contributions.}
\csname @beginparpenalty\endcsname=10000
\begin{itemize}
  \item \textbf{Corpus and benchmark.} We merge ResPlan and MagicPlan into one corpus of 73{,}581 plans with wall, door and passage labels, and score generators by brief rarity.
  \item \textbf{\method.} A masked graph generator, trained on real plans alone, that generates room graphs meeting a user's brief. At every step it is told which requirements are met, open or dead. For monotone requirements this status is exact, and standard message-passing encoders provably cannot infer it in general (\cref{sec:method}).
  \item \textbf{Results.} Without a certifier, the status lifts success on hard briefs from 0.105 to 0.467, while a more expressive encoder without it stays at 0.112. With a certifier, every returned graph meets the brief, and HouseDiffusion turns these graphs into floorplans that meet nearly twice as many briefs as with unconstrained sampling (\cref{sec:downstream}).
\end{itemize}
\section{Related work}
\label{sec:related}

\paragraph{Floorplans from graphs and graphs for floorplans.}
Most learned floorplan generators draw geometry from a room graph or from user constraints, including House-GAN++~\cite{nauata2021houseganpp}, Graph2Plan~\cite{hu2020graph2plan}, HouseDiffusion~\cite{shabani2023housediffusion}, later diffusion models~\cite{hu2025gsdiff} and language models trained with verifiable rewards~\cite{lara2026generative}. Fewer generate the graph itself or bypass it. Merrell \etal sample adjacencies from a Bayesian network learned from real plans~\cite{merrell2010computer}, BubbleFormer generates bubble diagrams in a given boundary~\cite{sun2023bubbleformer}, Xia \etal grow user-seeded graphs with DGMG~\cite{xia2024interactive}, MaskPLAN completes partial plans~\cite{zhang2024maskplan}, and language models map text or constraints to layouts~\cite{leng2023tell2design,qin2024chathousediffusion,luo2024dstruct2design}. None of these certifies that its output meets the user's requirements and validity, or reports results by how often the data supports a request.

\paragraph{Constrained graph and layout generation.}
Autoregressive~\cite{you2018graphrnn,li2018learning} and diffusion~\cite{vignac2023digress} graph generators fit the data without hard constraints. Conditioned graph generators further steer sampling toward target node or global attributes, for example via feature-conditioned latent diffusion~\cite{evdaimon2024neural}.
Such constraints are enforced by masking invalid actions, as for molecular valency~\cite{liu2018cgvae}, or by projecting the diffusion state onto a constraint set~\cite{madeira2024construct,sharma2024prodigy}, as also done for discrete sequences~\cite{cardei2025constrained}. NSPSG projects with an SMT solver inside the reverse process~\cite{zhang2026nspsg}, and NSGGM uses one to assemble learned proposals into certified molecules~\cite{geng2026nsggm}. Layout models constrain decoding~\cite{jiang2023layoutformerpp,inoue2023layoutdm}. All of these act on the generator's state or output, and none tells the network whether its partial graph can still satisfy a requirement.

\paragraph{Feasibility in sequential generation.}
Controlled text generation learns the probability that a prefix's completion satisfies a constraint and reweights the base model~\cite{yang2021fudge,meng2022nado}, or computes it exactly under a tractable surrogate for lexical or regular constraints~\cite{zhang2023gelato,zhang2024ctrlg}. Completion engines mask tokens after which no valid program can follow~\cite{poesia2022synchromesh}, shields override actions that would violate a temporal specification~\cite{alshiekh2018shielding}, and sequential Monte Carlo steers a model with prefix potentials and importance weights~\cite{lew2023smc,zhao2024twisted,loula2025syntactic}, since masking alone distorts the conditional~\cite{park2024gad}. In this tradition, our guard is a sound mask with exact weights for global graph requirements such as connectivity and reachability. Mention Flags feed a text decoder whether each required word has appeared yet~\cite{wang2021mention}, but cannot mark a constraint as impossible. Program synthesis conditions on the execution state of a partial program~\cite{chen2019execution,ellis2019write}, and hindsight relabeling trains on the goals an outcome met~\cite{andrychowicz2017her,zhang2023hir}. To our knowledge, no prior work trains a graph generator by hindsight on real data while feeding it the three-valued status of global requirements and the outcomes each undecided pair can still take, or shows that this one input reduces the work of several exact certifiers at once.
\section{A homogenized floorplan topology corpus}
\label{sec:data}

\paragraph{Sources and rooms.}
ResPlan~\cite{abouagour2025resplan} and the MagicPlan corpus released with PuzzleFusion~\cite{hosseini2023puzzlefusion} differ in room labels, units and door records, so we homogenize their vector plans into one schema with shared room classes and relation rules. Every room maps to one of 11 classes $\mathcal{C}$ (living, bedroom, kitchen, dining, bathroom, circulation, study, storage, utility, balcony and stair), and plans with an unmappable or invalid room are excluded. This keeps 9{,}365 of 17{,}000 ResPlan and 64{,}216 of 77{,}756 MagicPlan records, a total of 73{,}581 plans with 446{,}829 rooms (\cref{sec:supp-data}).

\paragraph{Adjacency and access.}
Two rooms are adjacent (they share a wall) when facing, nearly parallel boundary segments lie at most a wall's thickness apart and overlap long enough outside any third room. Door and passage records are matched to these walls, and a pair has access when it has a door or a passage. A pair with no recorded opening has none, but if a nearby record is unmatched, the affected pairs are left unlabeled (4\% of ResPlan and 41\% of MagicPlan training access pairs).

\paragraph{Splits and evaluation cohort.}
Exact duplicates under rotation and reflection are removed, and the variants of one geometry are hashed as one family into an 80/10/10 split (58{,}966, 7{,}414 and 7{,}201 plans). As briefs need door and passage labels, we evaluate on the 1{,}988 complete test programs (822 ResPlan, 1{,}166 MagicPlan), which have no unlabeled door or passage pair and whose room list matches at least one complete training plan.
\section{Problem: room graphs under a brief}
\label{sec:problem}
\paragraph{Programs and room graphs.}
A program $\prog=(c_1,\dots,c_n)$ with $c_i\in\mathcal C$ and $n\le17$ lists the rooms of a plan in no meaningful order. The output is a typed room graph $\G=(V,o)$ on the rooms $V=\{1,\dots,n\}$, in which every pair $e=\{i,j\}$ of $E=\binom V2$ takes one of five outcomes $o_e\in\Sigma=\{$\textsc{none}, \textsc{wall}, \textsc{wall+door}, \textsc{wall+passage}, \textsc{wall+door+passage}$\}$, so an opening always needs a shared wall. Its wall graph holds the pairs with a wall, its access graph those with a door or passage. Generation is thus edge-label prediction on typed nodes. Relabeling identical rooms gives the same plan, so models must be permutation equivariant and are evaluated up to relabeling. Accounting for permutation symmetries is a core requirement in deep graph generative modeling~\cite{bresson2026graviti}.

\paragraph{Requirements and task.}
A requirement is a predicate $R(\G)\in\{0,1\}$ on room graphs. Validity asks that the access graph be connected, that shared rooms connect without crossing a private room (bathroom, bedroom, storage or balcony), that every private room be reachable from a shared room via at most one other private room, and that the wall graph plus an exterior node joined to every bedroom and living room be planar. Real plans pass all four 96\% of the time. A user brief $\spec$ is a conjunction of possibly negated atoms of seven kinds, namely pair relations, type counts, room and type degrees, distances, reachability avoiding a room type and walk-through-free rooms (\cref{sec:supp-reqs}). With $\mathcal R$ the validity requirements and the atoms of $\spec$, a graph is \emph{certified}, written $\G\models\mathcal R$, when $R(\G)=1$ for every $R\in\mathcal R$. Given $\prog$ and $\spec$, the task is to return certified graphs close to real plans, ideally samples of $p(\G\mid\prog,\G\models\mathcal R)$ for a model $p$ of real plans, or report the brief infeasible.

\paragraph{The brief and its support.}
Our evaluation brief applies six requests wherever the program allows them, namely (B1) no bedroom door to the kitchen, (B2) at least one en-suite bedroom--bathroom door, (B3) no walk-through bedroom, (B4) at least three openings in the living room, (B5) a kitchen reachable from the living room without passing a bedroom, and (B6) no bathroom opening into the kitchen. This brief is never used in training. The \emph{support} of a brief is the share of complete training plans with the same program that meet it, grouped into four strata (none, below 5\%, 5--50\%, at least 50\%). \emph{Hard} briefs are those of the 584 MagicPlan test programs with support below 5\%, and 97 of them cannot be met, so no method can exceed 0.834 there. ResPlan has only 10 such programs, and a second, held-out brief tests its rare briefs (\cref{sec:exp-ablation}).

\begin{figure}[t]
  \centering
  \includegraphics{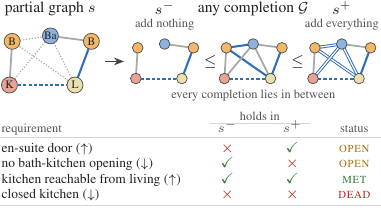}
  \caption{\textbf{Status from two extreme completions.} For $s$ of \cref{fig:method}, $s^-$ adds nothing and $s^+$ adds everything. Every completion lies between them, so a requirement that added structure never hurts ($\uparrow$) is \textsc{met} if it holds in $s^-$ and \textsc{dead} if it fails in $s^+$, with roles swapped for $\downarrow$ (\cref{lem:status}).}
  \label{fig:status}
\end{figure}

\paragraph{Partial graphs and status.}
A generator builds the graph pair by pair, so it holds a partial graph $s$ in which some pairs are undecided, and a completion of $s$ decides them all. We order graphs by inclusion of walls, doors and passages. A requirement is \emph{monotone up} if adding structure never breaks it, as for connectivity, reachability and minimum counts, and \emph{monotone down} if adding structure never helps, as for bans, count caps and planarity. Every validity requirement and every atom of our briefs is one of the two or reduces to them (\cref{sec:supp-status}). Let $s^-$ set every undecided pair to \textsc{none}, let $s^+$ set it to its largest outcome still consistent with $s$, and let $s[e{\leftarrow}o]$ decide pair $e$ as $o$. For a monotone-up $R$ the status of $R$ at $s$ is
\begin{equation}
  \mathrm{st}_R(s)=\begin{cases}\textsc{met} & R(s^-)=1,\\ \textsc{dead} & R(s^+)=0,\\ \textsc{open} & \text{otherwise},\end{cases}
  \label{eq:status}
\end{equation}
with $s^-$ and $s^+$ swapped for monotone-down $R$.

\begin{lemma}[Exact status]
\label{lem:status}
For monotone-up $R$, $R$ holds in every completion of $s$ iff $R(s^-)=1$, and in some completion iff $R(s^+)=1$ (swap for monotone-down).
\end{lemma}
\begin{proof}
Every completion $\G$ of $s$ satisfies $s^-\le\G\le s^+$, and both are completions, so $R(s^-)\le R(\G)\le R(s^+)$.
\end{proof}
An exhaustive check against all completions of small partial graphs found no mismatch in 8{,}640 cases. Conjunctions follow Kleene's three-valued logic~\cite{kleene1952metamath}. Like approximations of certain answers in incomplete databases~\cite{imielinski1984incomplete,libkin2016sql}, this is sound but not complete, since two open requirements may be jointly unsatisfiable. Hence a brief with a requirement \textsc{dead} on the empty partial graph is flagged infeasible before any sampling.
\section{Status-guided generation}
\label{sec:method}

\paragraph{Backbone.}
A masked edge generator~\cite{chang2022maskgit} draws an ordered partition $C=(C_1,\dots,C_T)$ of $E$ uniformly at random rather than by confidence, with cosine-scheduled sizes ($T{=}16$). Unlike confidence- or progress-aware unmasking schedules in discrete masked diffusion~\cite{mohamed2026fast}, random partitioning decouples step ordering from model certainty when hard feasibility constraints are enforced. Step $t$ decides the pairs of $C_t$,
\begin{equation}
  p_\theta(\G\mid C)=\prod_{t=1}^{T}\prod_{e\in C_t}q_\theta\big(o_e\mid s_{t-1},\prog,\spec\big),
  \label{eq:backbone}
\end{equation}
where $s_0$ is the empty partial graph and $s_t=s_{t-1}\cup\{o_e\}_{e\in C_t}$, with $p_\theta(\G)=\mathbb E_C\,p_\theta(\G\mid C)$. Each $q_\theta(\cdot\mid s,\prog,\spec)$ is a distribution over $\Sigma$ factorized as $P(\mathrm{wall})\,P(\mathrm{door}\mid\mathrm{wall})\,P(\mathrm{passage}\mid\mathrm{wall})$, computed by a symmetric pair head on a Transformer~\cite{vaswani2017attention} over the rooms without positional encodings. An adapter adds an embedding of every atom of the brief to the rooms and pairs it names. The backbone is trained by masked likelihood on real plans with hindsight briefs, whose atoms are drawn from a template given the program and take their polarity from the plan, so that the plan satisfies its own brief~\cite{andrychowicz2017her,zhang2023hir}.

\subsection{The status model}
\label{sec:method-status}
Whether a requirement such as connectivity survives the next decision is a global question, since closing a pair kills connectivity exactly when it is a bridge of the access graph of $s^+$. Our encoder cannot see this. \Cref{sec:supp-status} gives two partial graphs that 1-WL-bounded message passing cannot separate~\cite{zhang2023biconnectivity}, so they get identical room embeddings and pair scores although only one has such a bridge. We therefore supply the status as an input, and \cref{sec:exp-ablation} shows that a more expressive encoder does not make up for it. This aligns with theoretical analyses showing that standard message-passing networks and graph transformers face strict expressive limits when resolving global combinatorial properties such as connectivity and shortest-path reachability without explicit structural encoding~\cite{nikolentzos2026theoretical}.

\paragraph{Input.}
For a partial graph $s$, the \emph{status oracle} returns the status of every requirement in force and, for every undecided pair $e$, the outcomes that keep every requirement alive,
\begin{equation}
  \begin{gathered}
  \mathrm{st}(s)=\Big(\big(\mathrm{st}_R(s)\big)_{R\in\mathcal R},\ \big(\hat{\mathcal A}_e(s)\big)_{e}\Big),\\
  \mathcal A_e(s)=\big\{o\in\Sigma:\ \mathrm{st}_R(s[e{\leftarrow}o])\neq\textsc{dead}\ \ \forall R\in\mathcal R\big\},
  \end{gathered}
  \label{eq:allowed}
\end{equation}
where $\hat{\mathcal A}_e(s)\supseteq\mathcal A_e(s)$ is a cheaper over-approximation that tests each outcome on $s^-$ against the monotone-down requirements only. Each atom's embedding receives its \textsc{met} and \textsc{dead} flags, four added embeddings carry those of the validity requirements, and each undecided pair's embedding receives its five allowed bits. The status model $q_\theta(o_e\mid s,\prog,\spec,\mathrm{st}(s))$ replaces $q_\theta$ in \cref{eq:backbone} with the same encoder, heads and loss.

\paragraph{Hindsight training.}
For a training plan $\G^\star$ with a hindsight brief $\spec^\star$, we hide a random cosine-rate subset $M\subseteq E$ of its pairs, giving the partial graph $s_M$, and minimize
\begin{equation}
  \mathcal L(\theta)=-\,\mathbb E_{\G^\star,\spec^\star,M}\sum_{e\in M}\log q_\theta\big(o^\star_e\mid s_M,\prog,\spec^\star,\mathrm{st}^\star(s_M)\big),
  \label{eq:hindsight}
\end{equation}
where $\mathrm{st}^\star$ is the status of the requirements $\mathcal R^\star$ that $\G^\star$ meets, namely the validity requirements it passes and the atoms of $\spec^\star$.
\begin{lemma}[Hindsight consistency]
\label{lem:hindsight}
No requirement of $\mathcal R^\star$ is \textup{\textsc{dead}} at $s_M$, and $o^\star_e\in\mathcal A_e(s_M)$ for every $e\in M$.
\end{lemma}
\begin{proof}
$\G^\star$ completes both $s_M$ and $s_M[e{\leftarrow}o^\star_e]$ and meets every $R\in\mathcal R^\star$, so by \cref{lem:status} no $R$ is \textsc{dead} at either.
\end{proof}
The targets are therefore always allowed outcomes, so \cref{eq:hindsight} teaches the model which pairs must still carry which outcome without any reward or solver.

\subsection{Certifiers}
\label{sec:method-cert}
The status model raises the chance that a draw meets the brief but guarantees nothing. A \emph{certifier} turns draws into certified graphs, and all three below are exact and apply unchanged to both the status model and the plain backbone.

\paragraph{Status guard.}
Inside step $t$ the guard decides the pairs of $C_t$ one after another and writes each outcome into the current partial graph $s$. For pair $e$ it samples the model's step-$t$ distribution restricted to the exact allowed set,
\begin{equation}
  q_{\mathrm g}(o\mid s)=\frac{q_\theta(o\mid s_{t-1},\cdot)\,\mathbf 1[o\in\mathcal A_e(s)]}{Z_e(s)},
  \label{eq:guard}
\end{equation}
where $Z_e(s)=\sum_{o'\in\mathcal A_e(s)}q_\theta(o'\mid s_{t-1},\cdot)$ is the mass the guard keeps. If $\mathcal A_e(s)$ is empty the guard abstains, and if a requirement is already \textsc{dead} at $s_0$ it reports the brief infeasible.

\begin{proposition}[Certainty and exact weights]
\label{prop:certainty}
(i) Every graph the guard returns satisfies $\G\models\mathcal R$. (ii) With $w(C,\G)=\prod_{t}\prod_{e\in C_t}Z_e(s)$ (and $w=0$ on abstention), self-normalized $w$-weighted guarded samples estimate expectations under $p_\theta(\G\mid\G\models\mathcal R)\propto p_\theta(\G)\,\mathbf 1[\G\models\mathcal R]$.
\end{proposition}
\begin{proof}
(i) At the end no pair is undecided, so $s^-=s^+=\G$ and ``not \textsc{dead}'' means $R(\G)=1$ (\cref{lem:status}). (ii) By \cref{eq:backbone,eq:guard}, $p_\theta(\G\mid C)/q_{\mathrm g}(\G\mid C)=w(C,\G)$ for every returned $\G$, and $C$ has the same law under both. Every $\G\models\mathcal R$ with $p_\theta(\G\mid C)>0$ has $q_{\mathrm g}(\G\mid C)>0$, because each of its outcomes leaves $\G$ itself as a completion meeting $\mathcal R$ and is therefore allowed (\cref{lem:status}). Importance sampling over $(C,\G)$ then targets $p_\theta(C,\G\mid\G\models\mathcal R)$, whose $\G$-marginal is the stated conditional.
\end{proof}

\paragraph{Exact minimum-edit repair.}
A complete draw $\hat\G$ is replaced by
\begin{equation}
  \G_{\mathrm{rep}}=\arg\min_{\G\models\mathcal R}\ \sum_{e\in E}\mathbf 1\big[o_e(\G)\neq o_e(\hat\G)\big],
  \label{eq:repair}
\end{equation}
the nearest certified graph in changed pairs. We solve it by SMT projection as in NSPSG~\cite{zhang2026nspsg}, encoding our requirements for Z3~\cite{demoura2008z3} with lazy cuts for connectivity, planarity and path atoms. The solver has a limit of 3\,s per graph. When the limit runs out (0.01--0.42\% of draws) or Z3 proves the brief unsatisfiable, the draw is returned unrepaired and counts as a failure. Repair is deterministic given the draw, so its outputs cannot be reweighted.

\paragraph{Generate-and-filter.}
The model draws until $\G\models\mathcal R$, for at most 64 draws. This is exact rejection sampling from $p_\theta(\G\mid\G\models\mathcal R)$ whenever it returns, at an expected cost of $1/p_\theta(\G\models\mathcal R)$ draws.

\section{Experiments}
\label{sec:exp}

\paragraph{Setup.}
All methods are evaluated on the same 1{,}988 complete test programs, with 8 draws per program for each of three generator seeds. The brief suite applies the brief of \cref{sec:problem}, which 97 programs cannot meet, so success is at most 0.951. The mined suite gives each program the hindsight brief of its own real plan, as in training (\cref{sec:method}), and is feasible for all but 2 programs. Intervals are 95\% bootstrap intervals that resample programs with all their draws, and differences are paired on the same programs.

\paragraph{Metrics.}
A draw succeeds when it is returned, passes every validity requirement and meets every atom, so abstentions count as failures. Abstention is the share of draws a certifier gives up on although the brief is feasible. Bias is how far a certifier pushes graphs from the model's own conditional $p_\theta(\G\mid\G\models\mathcal R)$. It is measured on hard briefs as extra access edges per graph against exact rejection samples from it, so 0 is ideal. Realism is the maximum mean discrepancy (MMD$^2$) to real test graphs in access degree, wall clustering and wall spectrum (acc~/~clus~/~spec)~\cite{you2018graphrnn,liao2019gran}, reported $\times10^3$ only where the brief has at least 5\% support and never on hard briefs, since rarer briefs ask for what the real plan lacks. For reference, real training graphs score 0.1~/~1.4~/~1.2. Cost is single-core CPU seconds per graph.

\subsection{The status input under every certifier}
\label{sec:exp-main}

\begin{table*}[t]
  \centering
  \small
  \caption{\textbf{One status input, three exact certifiers.} The same backbone with (status) and without (plain) the status input, and hard denotes the 584 hard briefs (ceiling 0.834). The guard's importance weights reduce its bias to $-0.002$ (status) and $+0.026$ (plain).}
  \label{tab:certifiers}
  \resizebox{\textwidth}{!}{%
\begin{tabular}{@{}llcccccc@{}}
\toprule
Certifier & Model & Success (all) $\uparrow$ & Success (hard) $\uparrow$ & Abst. $\downarrow$ & Bias (hard) $\to0$ & MMD$\times10^3$ acc / clus / spec $\downarrow$ & CPU s $\downarrow$ \\
\midrule
\multirow{2}{*}{Sampling only} & plain & 0.422 & 0.105\,{\scriptsize[0.095, 0.115]} & 0.000 & -- & 4.4 / 9.7 / 3.0 & 0.011 \\
 & status & 0.675 & 0.467\,{\scriptsize[0.444, 0.491]} & 0.000 & -- & 6.6 / 6.0 / 2.8 & 0.15 \\
\addlinespace[2pt]
\multirow{2}{*}{Generate-and-filter ($\le$64)} & plain & 0.921 & 0.732\,{\scriptsize[0.700, 0.764]} & 0.032 & exact & 7.8 / 10.0 / 2.5 & 0.063 \\
 & status & 0.948 & 0.822\,{\scriptsize[0.791, 0.852]} & 0.004 & exact & 9.0 / 6.1 / 2.6 & 0.31 \\
\addlinespace[2pt]
\multirow{2}{*}{Exact min-edit repair} & plain & 0.950 & 0.833\,{\scriptsize[0.803, 0.863]} & 0.000 & +0.74\,{\scriptsize[+0.69, +0.80]} & 18.0 / 6.1 / 2.2 & 0.051 \\
 & status & 0.950 & 0.833\,{\scriptsize[0.803, 0.863]} & 0.000 & +0.17\,{\scriptsize[+0.15, +0.19]} & 11.0 / 5.4 / 2.7 & 0.17 \\
\addlinespace[2pt]
\multirow{2}{*}{Guard} & plain & 0.925 & 0.799\,{\scriptsize[0.769, 0.828]} & 0.028 & +0.52\,{\scriptsize[+0.48, +0.56]} & 15.2 / 7.3 / 2.5 & 0.22 \\
 & status & 0.921 & 0.798\,{\scriptsize[0.768, 0.827]} & 0.032 & +0.13\,{\scriptsize[+0.12, +0.15]} & 11.1 / 6.1 / 2.4 & 0.34 \\
\bottomrule
\end{tabular} }
\end{table*}

Without a certifier, the status input lifts hard-brief success from 0.105 to 0.467 (per seed 0.38--0.56). The gain comes from meeting the brief, as on hard briefs the share of draws meeting it rises from 0.15 to 0.53 while validity stays at 0.85 and 0.84. Realism moves in both directions (\cref{tab:certifiers}). With a certifier every returned graph meets the brief, so the gain shows up as less correction. Generate-and-filter needs 2.9--4.0 status draws per sample on feasible hard briefs instead of 15--19 plain draws, so within 64 draws hard-brief success rises from 0.732 to 0.822 and abstention there falls from 0.122 to 0.014. Exact repair reaches the ceiling with either model, but only 27\% of status draws need repair against 53\% of plain draws, so the status model needs 45\% fewer edits (0.38 against 0.70 changed pairs per graph). Its repaired graphs also stay closer to real access structure (\cref{tab:certifiers}). The guard succeeds equally often with and without the status input, since only abstention fails a guarded draw, but intervenes far less, as its effective sample size rises from 28\% to 80\% of the draws and the bias falls from $+0.52$ to $+0.13$ extra access edges. But, the status oracle raises sampling cost from 0.011 to 0.15\,s per graph.

\paragraph{Where the status helps.}
Without a certifier, the status input raises success by 0.31--0.49 when under 50\% of the training plans with the same rooms meet the brief, and by only 0.11 otherwise (\cref{tab:strata}), as the backbone's prior already favors such briefs. Under filtering, the gain is concentrated where no such plan meets the brief (success 0.639 for plain, 0.749 for status), as the rest are near the ceiling.

\subsection{Ablations and controls}
\label{sec:exp-ablation}

\paragraph{The status or the encoder?}
PPGN pair blocks~\cite{maron2019provably} lift our encoder beyond the 1-WL bound of \cref{sec:method-status}. Adding them does not help on hard briefs (success 0.112 without the status input, against 0.105 for our encoder), while the status input lifts PPGN to 0.503 (\cref{tab:supp-encoder}). The gain comes from the information, not encoder capacity.

\paragraph{A held-out brief.}
To test whether the gain depends on our six requests, we fixed a second brief with other requests before any evaluation (\cref{tab:supp-heldout}). On MagicPlan the gain recurs where no training plan meets the brief (sampling 0.08 to 0.20, filtering 0.37 to 0.72). On ResPlan, where no program has support for it, sampling stays near zero with or without the status input and only certifiers meet the brief.

\paragraph{Sampler variants.}
Sequential Monte Carlo~\cite{lew2023smc} resamples guarded draws by their importance weights after each step. This changes little on hard briefs (success $+0.003$ plain and $+0.005$ status), as only abstention fails a guarded draw. GRPO fine-tuning~\cite{shao2024deepseekmath} with the guard gives the highest success among our sampling methods (0.827 against 0.798 for status + guard) but the worst realism and fewer distinct graphs among a program's 8 draws (5.3 against 6.8), as reward fine-tuning narrows the model toward graphs that satisfy the brief.

\subsection{Comparison with published generators and certifiers}
\label{sec:exp-published}

\begin{table*}[t]
  \centering
  \small
  \caption{\textbf{Published generators}, alone and with a certifier, conditioned on the brief (\cref{sec:supp-methods}). $^\dagger$Reduced training budget (DGMG 6 epochs, Xia \etal 10 of 100 epochs), so these are lower bounds. Retrieve-and-filter returns matching training plans that meet the brief.}
  \label{tab:published}
  \resizebox{\textwidth}{!}{%
\begin{tabular}{@{}lcccccc@{}}
\toprule
Method & Success (all) $\uparrow$ & Success (hard) $\uparrow$ & Mined $\uparrow$ & Abst. $\downarrow$ & MMD$\times10^3$ acc / clus / spec $\downarrow$ & CPU s $\downarrow$ \\
\midrule
GraphRNN~\cite{you2018graphrnn} & 0.371 & 0.024 & 0.592 & 0.000 & 0.1 / 3.5 / 2.7 & 0.000 \\
\quad + guard & 0.938 & 0.803 & 0.956 & 0.014 & 12.9 / 3.1 / 2.9 & 0.24 \\
\quad + exact repair & 0.951 & 0.834 & 0.997 & 0.000 & 22.1 / 1.6 / 3.5 & -- \\
DGMG~\cite{li2018learning}$^\dagger$ & 0.125 & 0.052 & 0.313 & 0.000 & 16.6 / 5.2 / 2.1 & 0.009 \\
\quad + guard & 0.889 & 0.781 & 0.876 & 0.065 & 76.0 / 22.3 / 11.3 & 0.53 \\
\quad + exact repair & 0.950 & 0.833 & 0.995 & 0.000 & 55.5 / 2.9 / 1.9 & -- \\
Xia \etal~\cite{xia2024interactive}$^\dagger$ & 0.192 & 0.037 & 0.405 & 0.000 & 2.8 / 16.6 / 3.0 & -- \\
\quad + guard & 0.335 & 0.080 & 0.666 & 0.648 & 1.6 / 11.4 / 3.1 & -- \\
\quad + exact repair & 0.949 & 0.833 & 0.995 & 0.000 & 26.6 / 6.2 / 3.2 & -- \\
DiGress~\cite{vignac2023digress} & 0.675 & 0.268 & 0.820 & 0.000 & 3.7 / 0.7 / 0.5 & 2.77 \\
\quad + exact repair & 0.951 & 0.834 & 0.995 & 0.000 & 10.6 / 0.8 / 1.0 & -- \\
ConStruct~\cite{madeira2024construct} & 0.512 & 0.060 & 0.753 & 0.000 & 0.7 / 1.5 / 1.1 & 5.79 \\
NSPSG~\cite{zhang2026nspsg} & 0.951 & 0.834 & 0.997 & 0.000 & 23.2 / 1.5 / 1.4 & 5.83 \\
MILP 8-best (plain model) & 0.944 & 0.809 & 0.994 & 0.008 & 27.6 / 64.0 / 47.5 & 0.016 \\
Retrieve-and-filter & 0.808 & 0.370 & 0.951 & 0.151 & 6.1 / 0.8 / 0.6 & -- \\
GRPO + guard & 0.949 & 0.827 & 0.982 & 0.002 & 25.3 / 25.7 / 10.0 & 0.23 \\
\textbf{Ours: status + repair} & 0.950 & 0.833 & 0.996 & 0.000 & 11.0 / 5.4 / 2.7 & 0.17 \\
\textbf{Ours: status + guard} & 0.921 & 0.798 & 0.955 & 0.032 & 11.1 / 6.1 / 2.4 & 0.34 \\
\bottomrule
\end{tabular} }
\end{table*}

\begin{figure*}[t]
  \centering
  \includegraphics[width=0.92\textwidth]{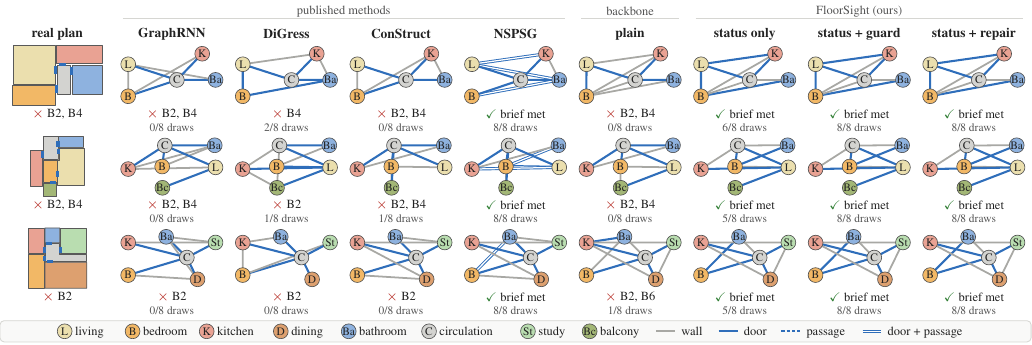}
  \caption{\textbf{Room graphs for rare briefs} (selected examples, see \cref{sec:supp-qual}). Each method's first returned graph for three MagicPlan programs whose brief no training plan meets, with the requests violated (B1--B6, \cref{sec:problem}) and the number of 8 draws that meet brief and validity. Without a certifier, graphs miss the en-suite door (B2) or the living-room openings (B4). The status input alone meets the brief in most draws here, and status + guard and status + repair give that same graph, while NSPSG ends with more openings than the real plan.}
  \label{fig:qual-graphs}
\end{figure*}

\paragraph{Generators without a certifier.}
The five published generators reach success 0.024--0.268 on hard briefs (\cref{tab:published}), far below the 0.834 that can be met. Their draws there are mostly valid (0.64--0.93) but rarely meet the brief (0.03--0.34, \cref{fig:qual-graphs}). Real plans nearly always show validity but rarely a hard brief, so conditioned generators learn validity but not the brief, and retrieve-and-filter reaches only 0.370 for the same reason.

\paragraph{Exact certifiers distort differently.}
NSPSG's SMT projection inside ConStruct, exact repair and MILP decoding of the plain model's 8 most likely certified graphs (MILP 8-best) all reach or approach the feasibility ceiling, and repair certifies every published generator. At the ceiling every pipeline meets the brief, so realism and cost decide between them. Repair distorts a generator's access structure more, the further its draws are from the brief. DiGress with repair is the most realistic. Status + repair matches its access MMD (11.0 against 10.6) at 0.17\,s per graph instead of at least 2.77\,s, but not its wall structure, which DiGress samples have before repair. NSPSG doubles the access MMD of status + repair (23.2 against 11.0) at 5.8\,s, as ConStruct draws rarely meet the brief. MILP 8-best returns modes, not samples, and has by far the least realistic wall structure.

\begin{figure*}[t]
  \centering
  \includegraphics[width=0.92\textwidth]{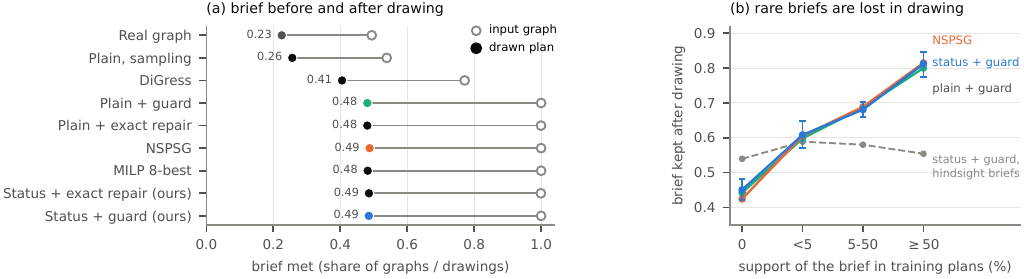}
  \caption{\textbf{What survives HouseDiffusion drawing} (HD seed 1, the 1{,}891 feasible briefs). (a) Brief met before and after drawing. (b) Brief kept for certified inputs by support stratum on MagicPlan (95\% intervals), with each program's own hindsight brief in gray.}
  \label{fig:downstream}
\end{figure*}

\begin{figure*}[t]
  \centering
  \includegraphics[width=0.92\textwidth]{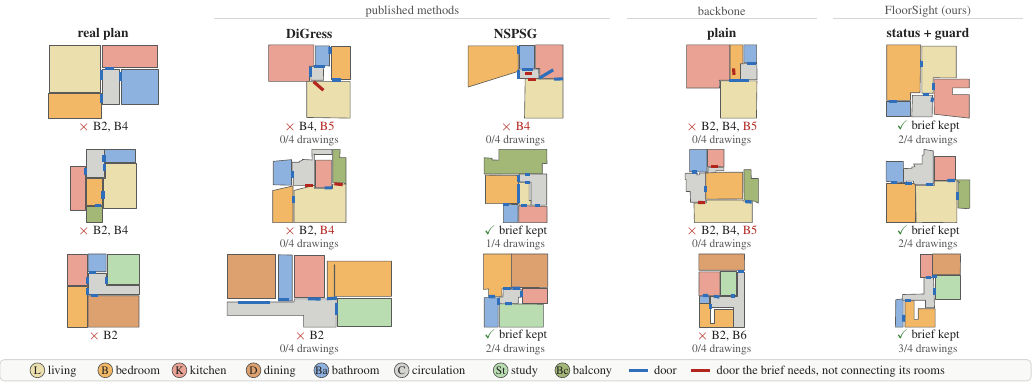}
  \caption{\textbf{Drawing the graphs of \cref{fig:qual-graphs}} (selected examples). Each graph is drawn four times, and each cell shows the first drawing keeping the brief, or else the first. Footers give the requests violated (B1--B6, \cref{sec:problem}), black if the graph misses them and red if drawing loses them, and how many of the four drawings keep it. Red bars are doors a lost request needs that the drawing does not connect to both rooms.}
  \label{fig:qual-drawings}
\end{figure*}

\paragraph{What the guard adds.}
The guard is neither the cheapest certifier nor the most successful. Repair on the same model succeeds more often at lower cost, and on the status model filtering reaches 0.822 against 0.798. Its distinct properties are exact importance weights (\cref{prop:certainty}), one draw per sample and plug-in use on sampled sequential generators, where it lifts GraphRNN from 0.024 to 0.803 and DGMG from 0.052 to 0.781 on hard briefs. It fails on the argmax decoder of Xia \etal, abstaining on 65\% of draws.
\section{What survives geometry?}
\label{sec:downstream}
\paragraph{Setup and metrics.}
Our guarantee holds for the room graph, so we test whether it survives drawing with HouseDiffusion (HD)~\cite{shabani2023housediffusion}. We retrain HD on our training split, changing only its graph interface. For each of the 1{,}891 programs with a feasible brief, the first graph of each method and generator seed is drawn $K{=}4$ times and read back into a room graph with the relation rules of \cref{sec:data}, which recover 99.7--100\% of the relations of real geometry. On each read-back graph we check the brief and validity again, and a drawing \emph{keeps} the brief if its read-back graph meets it, valid or not. We also report KID$\times10^3$~\cite{binkowski2018demystifying} of the rendered plans against real ones (lower is better, \cref{sec:supp-downstream}).

\paragraph{Certification survives in part.}
Certified graphs keep the brief in 0.46--0.50 of drawings, nearly double the 0.26 of plain sampling (\cref{fig:downstream}a). DiGress graphs keep it in 0.41 and real graphs in 0.23, as only half of real graphs meet it before drawing. All certifiers are within 0.005 of status + guard in single drawings, so what survives depends on the certificate, not on the certifier. Validity after drawing is 0.21--0.26 for every input, real graphs included, almost entirely because dropped openings disconnect the access graph (connected in 0.21--0.27 of drawings, planar in at least 0.98). Image realism follows graph realism (KID 58.5 for MILP, 46.3 for status + guard, 41.8 for real graphs).

\paragraph{Rare briefs are harder to draw.}
On MagicPlan, certified graphs keep the brief in 0.78--0.82 of drawings when at least half of the training plans with the same rooms meet it, but only in 0.41--0.49 when none does, whereas each program's own hindsight brief is kept equally often in every stratum (\cref{fig:downstream}b). In a linear probability model, this gap shrinks from $-0.33$ to $-0.09$ once we control for the brief's atoms and the input's openings and atypicality. The strongest predictor of a loss is the number of openings the brief rests on.

\paragraph{How geometry loses a requirement.}
Drawing keeps 61--64\% of a certified graph's openings, and keeps the openings a met request rests on barely more often than the others. Of the briefs lost with every room drawn, 89--92\% lose such an opening (\cref{fig:qual-drawings} and \cref{sec:supp-downstream}). HD never sees the brief, so it cannot know which openings the brief needs.

\paragraph{Screening drawings.}
Screening four drawings per graph with the brief check raises the share of programs with a passing plan from 0.47--0.50 with one drawing to 0.82--0.85 for certified graphs, against 0.43 for real graphs and 0.46 for plain sampling, no better than certified graphs drawn once. A separate run with sixteen drawings per graph reaches 0.98--0.99. On MagicPlan briefs that no training plan meets, status + guard and NSPSG graphs keep the brief equally often in one drawing ($+0.004$), but screening four drawings finds a passing plan more often from status + guard ($+0.084$ [0.051, 0.114]).
\section{Limitations and conclusion}
\label{sec:conclusion}

\paragraph{Limitations.}
Our generator produces only the room graph. Requirements on room sizes, shapes, windows or the boundary are outside its scope, and the plans are drawn by an existing layout generator that does not see the brief. The status is exact for monotone requirements only, and strongly interacting requirements need joint feasibility checks. When a brief goes against a corpus's conventions, as the held-out brief does on ResPlan, the learned gain vanishes and only the certifiers meet it.

\paragraph{Conclusion.}
We presented a methodology for generating floorplan room graphs that are valid and meet a user's brief. At every step, the exact feasibility status of each requirement is computed from two extreme completions. A masked generator takes this status as input and learns from real plans by hindsight, and exact certifiers guarantee every returned graph. The status lets the generator itself follow the brief. On rare briefs it lifts success from 0.105 to 0.467, while a more expressive encoder without it stays at 0.112, and it reduces the correction every certifier needs. Status + repair reaches the feasibility ceiling and matches the access MMD of the best certified published pipeline at a sixteenth of its cost. The certified graphs carry through to plans, nearly doubling the briefs kept by a diffusion-based layout generator, although drawing still loses about half of them. The natural next step is to generate topology and geometry jointly, so that floorplans keep their requirements from brief to plan.

\clearpage
\appendix
\renewcommand{\topfraction}{0.9}\renewcommand{\dbltopfraction}{0.9}
\renewcommand{\textfraction}{0.07}\renewcommand{\floatpagefraction}{0.8}
\renewcommand{\dblfloatpagefraction}{0.8}
\setcounter{topnumber}{3}\setcounter{dbltopnumber}{3}\setcounter{totalnumber}{5}

\noindent This appendix{} expands what the main paper states briefly. It covers how the corpus was built (\cref{sec:supp-data}), how validity and user briefs are modelled and given to the generator (\cref{sec:supp-reqs}), why the status is exact and why the encoder needs it (\cref{sec:supp-status}), and how the certifiers and published baselines are run (\cref{sec:supp-methods}). It then reports every graph metric with the ablations (\cref{sec:supp-results}), room graphs for random programs (\cref{sec:supp-qual}) and the downstream floorplans (\cref{sec:supp-downstream}).

\section{The homogenized corpus}
\label{sec:supp-data}

\paragraph{Why ResPlan and MagicPlan.}
RPLAN~\cite{wu2019data} is the training corpus of most learned floorplan generators~\cite{nauata2021houseganpp,hu2020graph2plan,shabani2023housediffusion}, but its creators no longer grant access to it, so we could not use it. We instead built our corpus from two public vector releases that record what our task needs, room polygons with room labels and door records: ResPlan~\cite{abouagour2025resplan} and the MagicPlan corpus released with PuzzleFusion~\cite{hosseini2023puzzlefusion}. Their design conventions differ. ResPlan has no corridor class, and its plans use the living room as the hub of circulation, while MagicPlan plans route through halls and corridors. This is why the support of a brief differs so much between the two (\cref{sec:problem}). Their room labels, units and door records also differ, so neither can be used as is, and we cleaned and homogenized both into one schema.

\paragraph{Rooms.}
Room polygons are validated (holes kept; containment and material overlap excluded), canonicalized and normalized by the room-union bounding box. No geometry is repaired, snapped or invented: a plan whose geometry fails a check is excluded with its reason (\cref{tab:supp-data}). Every room maps to one of 11 classes (\cref{tab:supp-classes}). A plan with a room that maps to no class, or with a garden, parking or pool layer (ResPlan), is excluded as a whole. No room is therefore dropped silently, and every graph keeps its complete program.

\begin{table}[!t]
  \centering\small
  \setlength{\tabcolsep}{3pt}
  \caption{\textbf{Room classes} and the native labels mapped to them ($^\ast$ a narrower label rolled up into the class). ResPlan has seven native room labels, MagicPlan nineteen.}
  \label{tab:supp-classes}
  \begin{tabular}{@{}lll@{}}
    \toprule
    Class & ResPlan & MagicPlan \\
    \midrule
    living & living & Living Room, Den$^\ast$ \\
    bedroom & bedroom & Bedroom, Master Bedroom$^\ast$ \\
    kitchen & kitchen & Kitchen \\
    dining & -- & Dining Room \\
    bathroom & bathroom & \makecell[l]{Bathroom, Master Bathroom$^\ast$,\\ Toilet$^\ast$, Wc$^\ast$, Restrooms$^\ast$} \\
    circulation & -- & Hall, Corridor \\
    study & -- & Study, Private Office$^\ast$ \\
    storage & storage & Storage, Closet$^\ast$ \\
    utility & -- & Laundry Room \\
    balcony & balcony & Balcony \\
    stair & stair & -- \\
    \bottomrule
  \end{tabular}
\end{table}

\paragraph{Wall adjacency.}
For every unordered room pair $\{a,b\}$ we compute the \emph{supported interface}

\begin{equation}
  \begin{aligned}
    I_{ab} ={}&
    \bigcup_{\substack{
      e\subset\partial R_a,\ e'\subset\partial R_b\\
      \angle(e,e')\le 5^\circ,\ \text{facing}
    }}
    \Big\{x\in\mathrm{proj}(e,e') :
      d(x)\le g_s\Big\}\\
    &\setminus \bigcup_{c\ne a,b}\mathrm{occ}_c,
  \end{aligned}
  \label{eq:interface}
\end{equation}

It collects the stretches where a boundary segment of room $a$ faces one of room $b$, within $5^\circ$ of parallel and at most $g_s$ apart along the whole stretch, minus any part covered by a third room $c$. The pair is adjacent iff some straight piece of $I_{ab}$ is at least $\ell_s$ long. ResPlan uses its per-plan wall depth for $g_s$ and $\ell_s=2g_s$; MagicPlan uses $g_s=0.5$ and $\ell_s=0.3$ source units. Corner contact never suffices. Every one of the $\binom{n}{2}$ pairs thus gets a definite adjacent or not-adjacent label.
The larger MagicPlan gap absorbs its thick or loosely scanned walls: with $g_s=0.15$, 18.0\% of MagicPlan plans would have a disconnected adjacency graph, against 1.7\% with 0.5.

\paragraph{Openings and unlabeled pairs.}
Door and passage records are matched to supported interfaces. A ResPlan connector polygon must lie on exactly one supported pair; ResPlan passages are long interface spans without a wall. A MagicPlan opening is recorded only by the room that owns it, without the second room, so a door counts only when the room on the other side records a matching opening on the same interface. Windows never give access. A pair with no recorded opening has no door or passage. When a nearby record could not be matched, however, the door and passage labels of the affected pairs are left unlabeled rather than set to absent. Under this policy, training access pairs are supervised on 191{,}823 of 200{,}018 ResPlan pairs (8{,}195 masked) and 486{,}384 of 827{,}707 MagicPlan pairs (341{,}323 masked); adjacency is never masked. Unlabeled pairs are left out of the door and passage terms of the training loss. The evaluation cohort keeps only programs with no unlabeled pair whose room list matches at least one complete training plan (\cref{sec:data}).

\paragraph{Deduplication and splits.}
Exact duplicates of geometry and labels under the eight rotations and reflections are removed. The remaining variants of one geometry (for example, the same rooms with different doors) form one family. Families are assigned to splits as a unit by hash, so no geometry appears in two splits (\cref{tab:supp-data}).

\begin{table}[!t]
  \centering\small
  \caption{\textbf{Corpus statistics} and the largest exclusion reasons. Float32 rows: geometry that changes when stored in 32-bit floats.}
  \label{tab:supp-data}
  \resizebox{\linewidth}{!}{%
  \begin{tabular}{@{}lr@{}}
    \toprule
    Quantity & Count \\
    \midrule
    Source records scanned (ResPlan / MagicPlan) & 17{,}000 / 77{,}756 \\
    Retained plans (ResPlan / MagicPlan)          & 9{,}365 / 64{,}216 \\
    Rooms / room pairs                             & 446{,}829 / 1{,}280{,}524 \\
    Train / val / test plans                       & 58{,}966 / 7{,}414 / 7{,}201 \\
    Evaluation cohort (ResPlan / MagicPlan)        & 822 / 1{,}166 \\
    \midrule
    \multicolumn{2}{@{}l}{\emph{Largest exclusions}} \\
    Material room overlap                          & 12{,}802 \\
    Float32 vertex collapse / adjacency change     & 2{,}888 / 1{,}871 \\
    Exact duplicate targets                        & 1{,}313 \\
    Room contained in another room                 & 709 \\
    Augmented copies without lineage (ResPlan)     & 683 \\
    Garden, parking or pool present (ResPlan) & 626 \\
    \bottomrule
  \end{tabular}}
\end{table}

\section{Validity, briefs and how they enter the model}
\label{sec:supp-reqs}

\paragraph{Validity.}
The four validity requirements of \cref{sec:problem} turn basic rules of residential design into exact predicates on the room graph (\cref{tab:supp-validity}). Private rooms $\V_{\mathrm p}$ are those of the classes bathroom, bedroom, storage, balcony; all other rooms are shared. $\cross(u,v)$ is the smallest number of private rooms one must pass through to go from $u$ to $v$ through openings (a shortest path in the access graph that counts private rooms). Shared transit is vacuous with fewer than two shared rooms, and private depth with no shared or no private room. The real test graphs of the 1{,}988 evaluation programs pass all four in 0.96 of programs (\cref{tab:supp-metrics}).

\begin{table}[!t]
  \centering\small
  \setlength{\tabcolsep}{3pt}
  \caption{\textbf{Validity requirements}, the design rule each encodes, and the predicate. Access within walls is built into the outcomes $\Sigma$.}
  \label{tab:supp-validity}
  \begin{tabular}{@{}p{1.65cm}p{2.35cm}p{3.95cm}@{}}
    \toprule
    Requirement & Design rule & Holds when \\
    \midrule
    Connected        & every room can be entered & the access graph is connected \\
    Shared transit   & moving between shared rooms crosses no private room & $\cross(u,v)=0$ for all shared rooms $u\ne v$ \\
    Private depth    & a private room opens onto a shared room, or onto a private room that does (en-suite, closet) & every private room $p$ has $\min_{s\notin\V_{\mathrm p}}\cross(s,p)\le1$ \\
    Exterior planar  & rooms tile the plane; bedrooms and living rooms have an outside wall & walls plus an exterior node joined to every bedroom and living room form a planar graph \\
    \bottomrule
  \end{tabular}
\end{table}

\paragraph{The constraint language.}
A brief is a conjunction of atoms. Each atom is one of seven kinds (\cref{tab:supp-atoms}), with a relation (wall, door, passage, access), a polarity and, where needed, bounds. Rooms have no names in the data, so an atom names a room by a \emph{handle}, its class and index in the program (\texttt{living\_1}). Type atoms apply to every room of a class. In a walk-through-free atom, annexes (bathroom, storage, balcony) may be reached through another room without counting as a walk-through. \Cref{tab:supp-briefs} lists the requests of the evaluation brief and of the held-out brief. A request is dropped when the program lacks its room types.

\begin{table}[!t]
  \centering\small
  \setlength{\tabcolsep}{3pt}
  \caption{\textbf{Atom kinds} of the constraint language. Each also carries a relation and a polarity.}
  \label{tab:supp-atoms}
  \begin{tabular}{@{}lp{5.6cm}@{}}
    \toprule
    Kind & Predicate \\
    \midrule
    pair             & rooms $a$, $b$ have the relation \\
    type count       & some pair of rooms of types $(T_1,T_2)$ has the relation \\
    degree           & room $a$ has a door (or access) count in $[lo, hi]$ \\
    type degree      & every room of type $T$ has a count in $[lo, hi]$ \\
    reach avoiding   & $b$ reachable from $a$ by access without entering type $F$ \\
    distance         & access hop distance from $a$ to $b$ at most $k$ \\
    walk-through-free & no room of type $T$ separates two non-annex rooms \\
    \bottomrule
  \end{tabular}
\end{table}

\begin{table}[!t]
  \centering\small
  \setlength{\tabcolsep}{3pt}
  \caption{\textbf{Evaluation brief (B) and held-out brief (H).} Direction: $\uparrow$ monotone up, $\downarrow$ monotone down, $\updownarrow$ an interval (floor $\wedge$ cap). The held-out brief was fixed before any evaluation. Of its six requests, three use an atom kind or relation that the evaluation brief does not (H1--H3), and three apply a known kind to other rooms or in the opposite direction (H4--H6).}
  \label{tab:supp-briefs}
  \begin{tabular}{@{}lp{4.15cm}lc@{}}
    \toprule
    & Request & Atom kind & Dir. \\
    \midrule
    B1 & no bedroom door to the kitchen & type count, door & $\downarrow$ \\
    B2 & an en-suite bedroom--bathroom door & type count, door & $\uparrow$ \\
    B3 & no walk-through bedroom & walk-through-free & $\uparrow$ \\
    B4 & living room has $\ge3$ openings & degree, access & $\uparrow$ \\
    B5 & kitchen reachable from the living room without passing a bedroom & reach avoiding & $\uparrow$ \\
    B6 & no bathroom opening into the kitchen & type count, access & $\downarrow$ \\
    \midrule
    H1 & every bedroom has exactly one door & type degree, door & $\updownarrow$ \\
    H2 & kitchen shares a wall with the living room & type count, wall & $\uparrow$ \\
    H3 & bathroom 1 at most two openings from bedroom 1 & distance & $\uparrow$ \\
    H4 & no bathroom opens onto the living room & type count, access & $\downarrow$ \\
    H5 & living room 1 has at most three openings & degree, access & $\downarrow$ \\
    H6 & bedroom 1 reachable from living room 1 without passing the kitchen & reach avoiding & $\uparrow$ \\
    \bottomrule
  \end{tabular}
\end{table}

\paragraph{From a brief to the model's input.}
\Cref{fig:supp-spec} follows one real brief through the pipeline. Each atom is stored as ten integers (kind, relation, polarity, rooms $a$ and $b$, types, avoided type and bounds, $-1$ when unused). The model encodes it as a feature vector (one-hot kind and relation, polarity, scaled bounds, one-hot named classes) together with its scope, the rooms it names and the room pairs it constrains. An adapter maps each atom to an embedding $e_k$ and adds it to three places: the rooms the atom names, the pairs it constrains, and, pooled over all atoms, every room. Scopes follow the room slots, so relabeling the rooms permutes the encoding and the model stays permutation equivariant.

The status model also receives the oracle's output. Each atom's vector gets its \textsc{met} and \textsc{dead} flags, and four extra tokens carry the flags of the validity requirements. Each undecided pair's state embedding (one of 27 codes, since its wall, door and passage are each masked, absent or present) receives a learned linear map of its five allowed bits. These bits are the cheaper over-approximation $\hat{\mathcal A}_e(s)\supseteq\mathcal A_e(s)$ of \cref{eq:allowed}: every outcome they exclude is truly dead. In \cref{fig:supp-spec}, half of the pairs are decided as in the real plan, and the bits equal $\mathcal A_e(s)$. The en-suite request B2 is still \textsc{open}, but the bedroom--bathroom pair is the last pair that can meet it, so only its door outcomes remain allowed. This is the pair-level information that the encoder cannot infer alone (\cref{sec:supp-status}).

\begin{figure*}[!t]
  \centering
  \begin{minipage}[t]{\textwidth}
    \footnotesize\textbf{(a)} The brief of the first program of \cref{fig:qual-graphs} as JSON, as built for evaluation:\par\vspace{-2pt}
    {\scriptsize%
\begin{verbatim}
{"rooms": [
  {"slot": 0, "type": "bathroom"},
  {"slot": 1, "type": "bedroom"},
  {"slot": 2, "type": "circulation"},
  {"slot": 3, "type": "kitchen", "handle": "kitchen_1"},
  {"slot": 4, "type": "living", "handle": "living_1"}],
 "constraints": [
  {"request": "B1", "kind": "type_count", "relation": "door", "types": ["bedroom", "kitchen"], "value": false},
  {"request": "B2", "kind": "type_count", "relation": "door", "types": ["bedroom", "bathroom"], "value": true},
  {"request": "B3", "kind": "walkthrough_free", "type": "bedroom", "value": true},
  {"request": "B4", "kind": "degree", "relation": "access", "room": "living_1", "min": 3, "max": null, "value": true},
  {"request": "B5", "kind": "reach_avoiding", "a": "living_1", "b": "kitchen_1", "avoid": "bedroom", "value": true},
  {"request": "B6", "kind": "type_count", "relation": "access", "types": ["bathroom", "kitchen"], "value": false}]}
\end{verbatim} }
  \end{minipage}\par\medskip
  \begin{minipage}[t]{0.56\textwidth}
    \footnotesize\textbf{(b)} Encoded atoms and their status on the empty graph $s_0$ and a partial graph $s$:\par\smallskip
    \resizebox{\linewidth}{!}{%
\begin{tabular}{@{}lccc@{}}
\toprule
Requirement & Encoded atom & $s_0$ & $s$ \\
\midrule
connected & -- & \textsc{open} & \textsc{open} \\
shared transit & -- & \textsc{open} & \textsc{met} \\
private depth & -- & \textsc{open} & \textsc{open} \\
exterior planar & -- & \textsc{open} & \textsc{met} \\
\midrule
B1 & (1, 1, 0, -1, -1, 1, 2, -1, -1, -1) & \textsc{open} & \textsc{met} \\
B2 & (1, 1, 1, -1, -1, 1, 4, -1, -1, -1) & \textsc{open} & \textsc{open} \\
B3 & (6, 3, 1, -1, -1, 1, -1, -1, -1, -1) & \textsc{open} & \textsc{open} \\
B4 & (2, 3, 1, 4, -1, -1, -1, -1, 3, -1) & \textsc{open} & \textsc{open} \\
B5 & (4, 3, 1, 4, 3, -1, -1, 1, -1, -1) & \textsc{open} & \textsc{met} \\
B6 & (1, 3, 0, -1, -1, 4, 2, -1, -1, -1) & \textsc{open} & \textsc{met} \\
\bottomrule
\end{tabular} }
  \end{minipage}\hfill
  \begin{minipage}[t]{0.40\textwidth}
    \footnotesize\textbf{(c)} The pairs of $s$ and the outcomes that keep every requirement alive:\par\smallskip
    \resizebox{\linewidth}{!}{%
\begin{tabular}{@{}lll@{}}
\toprule
Pair & In $s$ & Allowed outcomes $\mathcal A_e(s)$ \\
\midrule
Ba--C & wall+door & -- \\
Ba--K & wall & -- \\
B--K & none & -- \\
C--K & wall+door & -- \\
C--L & wall+door & -- \\
B--C & undecided & all five \\
Ba--L & undecided & all five \\
B--L & undecided & all five \\
K--L & undecided & all five \\
Ba--B & undecided & wall+door, wall+door+passage \\
\bottomrule
\end{tabular} }
  \end{minipage}
  \caption{\textbf{A real brief, its encoding and its status} (MagicPlan program with a bathroom Ba, a bedroom B, a circulation room C, a kitchen K and a living room L). Encoded fields: kind, relation, polarity, rooms $a$, $b$, types $T_1$, $T_2$, avoided type, $lo$, $hi$. The partial graph $s$ decides a seeded random half of the pairs as in the real plan, which misses B2 and B4. Computed with our oracle; here the model's allowed bits equal $\mathcal A_e(s)$.}
  \label{fig:supp-spec}
\end{figure*}

\paragraph{Hindsight briefs.}
Training briefs are mined from each training plan in two stages. First, the number of atoms (0 to 8) and every atom \emph{template} (kind, relation, rooms, types, thresholds) are drawn from the program alone, with handles bound to uniformly random rooms. Then the plan is read once per template, and each atom takes the polarity (or, for an exact degree, the count) that the plan has, so the plan always meets its own brief. Door, passage and access atoms need complete opening labels, so plans with unlabeled pairs get wall atoms only. Each training plan gets 8 mined briefs.
Because templates never look at the graph, brief $\spec$ is mined from plan $\G$ with probability $r(\spec\mid\G)=f(\spec\mid\prog)\,\mathbf 1[\G\models\spec]$, where $f$ is the template distribution. For a fixed $\spec$, the training plans paired with it therefore follow $p(\G\mid\prog)\,\mathbf 1[\G\models\spec]$ up to normalization. A model fit to these pairs by likelihood thus learns $p(\G\mid\prog,\G\models\spec)$, the target of \cref{sec:problem}, without any reward. Choosing a room because the plan has some property (say, the bedroom with the en-suite) would break this, which is why handles are bound before the plan is read.

\paragraph{Support.}
A program's support is estimated from at most 80 of its matching complete training plans (a seeded sample), with their rooms aligned to the program's by type order. The strata of \cref{sec:problem} then group programs by the share of these plans that meet the brief.

\section{The status: exactness and why it must be an input}
\label{sec:supp-status}

\paragraph{Every requirement is monotone or reduces to it.}
\Cref{tab:coverage} gives the direction of each validity requirement and atom kind. A count atom splits into a floor (up) and a cap (down), and its negation is the disjunction of the negated halves. Walk-through-free is up once the access graph is connected, so it is declared \textsc{dead} only in that case. Negated walk-through atoms are not monotone and are not guarded; they occur in mined briefs only (3.1\% of mined atoms), never in the evaluation briefs.
We checked every direction by brute force. We enumerated all completions of 160 random partial graphs (5--6 rooms, 3--6 undecided pairs) for the validity requirements and 23 atom variants: all 8{,}640 exists/for-all comparisons and all 56{,}376 single-pair veto comparisons agree with \cref{lem:status}. For a conjunction the status is sound but not complete: two requirements can each be met by some completion but not by the same one. Restricting the open completion by the bans already in force reduced such misses from 13\% to 2\% of the infeasible outcomes in that test, with no wrong veto. On the real test briefs, the infeasibility flag on the empty graph and exact Z3 reach the same verdict on every plan that Z3 resolves within 30\,s, and both flag the same 97 plans as infeasible.

\begin{table*}[!t]
\centering\small
\caption{\textbf{Coverage of the validity requirements and the seven atom kinds.} \emph{Open} completion: every undecided pair at its largest consistent outcome ($s^+$); \emph{closed}: every undecided pair \textsc{none} ($s^-$).}
\label{tab:coverage}
\begin{tabular}{@{}p{5.2cm}p{2.0cm}p{4.3cm}p{4.6cm}@{}}
\toprule
Requirement & Direction & \textsc{dead} when & Vetoed move\\
\midrule
Access connectivity; shared transit; private depth & up & open completion fails & closing a pair the open completion needs\\
Exterior planarity & down & placed walls already non-planar & a wall that breaks planarity\\
Access within adjacency & structural & -- & built into the outcomes\\
Pair, type count, degree floor, reach, distance (must hold) & up & open completion fails & closing the last candidate or route\\
Their negations; caps; bans & down & closed completion fails & opening past a cap or into a ban\\
Interval or exact count & floor $\wedge$ cap & either half dead & both rules\\
Walk-through-free & up, given connected & connected $\wedge$ it fails on the open completion & closing a pair that leaves a pass-through room\\
Negated walk-through & none & -- & not guarded\\
\bottomrule
\end{tabular}
\end{table*}

\paragraph{The status is invisible to the encoder.}
\begin{proposition}[Status is invisible to the encoders]
\label{prop:invisible}
Let a generator embed each room from its own features and the multiset of (neighbor features, pair state) over the other rooms, process rooms by any permutation-equivariant set function, and score a pair from its two room embeddings and its own state (our Transformer backbone, with pair states entering through one message round and the pair head, is of this form). Take ten rooms of one class. Let $B$ be the cubic graph made of two copies of $K_4$ with one edge subdivided, joined by an edge between the two subdivision vertices (one bridge), and $P$ the Petersen graph. Let $s_B$ ($s_P$) leave the 15 pairs of $B$ ($P$) undecided and decide all other pairs absent. Then the generator assigns identical distributions to all 15 undecided pairs in both $s_B$ and $s_P$, yet for ``access graph connected'' the bridge is the only critical pair of $s_B$ and $s_P$ has none.
\end{proposition}
\begin{proof}
In both partial graphs every room has three undecided and six absent incident pairs, so all room embeddings coincide and so do all pair scores. A 1-WL hash~\cite{shervashidze2011weisfeiler} confirms that $B$ and $P$ are indistinguishable, while $B$ has exactly one bridge and $P$ none; detecting bridges is beyond 1-WL-bounded message passing~\cite{zhang2023biconnectivity}.
\end{proof}
\noindent With the status input, the bridge of $s_B$ is the one pair whose \textsc{none} outcome is not allowed, so the two partial graphs receive different inputs.

\section{Certifiers, baselines and implementation}
\label{sec:supp-methods}

\paragraph{Certifiers.}
\emph{Exact repair} (\cref{eq:repair}) is a Z3 optimization with one Boolean per pair and outcome (exactly one true per pair), a soft clause per pair to keep its drawn outcome (so the optimum minimizes changed pairs) and every requirement as hard clauses, as in NSPSG~\cite{zhang2026nspsg}. Connectivity, planarity and path atoms are added lazily: the solver's graph is checked, and a cut is added for each violation (a Kuratowski subgraph for planarity). \emph{Generate-and-filter} draws until a draw is certified, at most 64 draws. \emph{MILP 8-best} returns the eight most likely graphs that meet every requirement, scored by the backbone's outcome probabilities with nothing decided (HiGHS~\cite{huangfu2018highs}, lazy cuts). \emph{Sequential Monte Carlo} resamples the guarded particles of a program by their importance weights after every step (systematic resampling). \emph{Retrieve-and-filter} returns training graphs with the same room program that meet the brief, searching up to 200 of them and up to 24 type-preserving room assignments.

\paragraph{Published generators.}
Every published model is run from its official repository at a pinned commit, trained on our training split with three seeds, and changed only where our task requires it. Each one receives the room program as fixed node types and outputs one of our five outcomes per pair. Each is conditioned on the brief through its own input interface, using the same atom encoding as our model, and trained on hindsight briefs like our backbone.
\begin{itemize}
  \item \textbf{GraphRNN}~\cite{you2018graphrnn} (commit 3444b8a, MIT): no file modified; BFS ordering; the edge-level output is a 5-way categorical over our outcomes, and node types and the spec enter each graph-level step.
  \item \textbf{DGMG}~\cite{li2018learning}: the DeepGMG model functions of the GraphRNN repository, used unchanged. Rooms are added with their own type (no add-node decision); a chosen edge receives one of the four non-\textsc{none} outcomes from a linear head. 6 epochs instead of the published schedule (CPU time), so a lower bound.
  \item \textbf{Xia \etal}~\cite{xia2024interactive} (no license; private comparison only): their housing DGMG with a DGL-free stand-in for message passing, no wall elements, 10 of 100 epochs. The brief enters their conditioning vector; positive pair atoms become their requested connections. Edge classes are the argmax of their feature head.
  \item \textbf{DiGress}~\cite{vignac2023digress} (commit bdfa10c, MIT) and \textbf{ConStruct}~\cite{madeira2024construct} (commit 72f329d, MIT): no edits to their code; all $T{=}500$ reverse steps. Their edge classes are our five outcomes, with \textsc{none} as the no-edge class; node types are fixed inputs and are not noised. The brief enters as atom counts and summed atom features in the global feature, per-room role counts and per-pair scope counts. ConStruct's own projection rejects every new edge that breaks a requirement that edge deletion preserves, namely exterior planarity and the down parts of the brief (bans, caps). It cannot express up requirements such as connectivity.
  \item \textbf{NSPSG}~\cite{zhang2026nspsg} (commit 73367b7, Apache-2.0): its procedure on our ConStruct models, SMT minimum-edit projection over the last 25 of 500 steps with our requirements in Z3 (3\,s per projection); its learned projector, trained per constraint to imitate the solver, is not used.
\end{itemize}
For GraphRNN, DGMG and Xia \etal the plug-in guard restricts each decision's distribution to outcomes that leave no requirement dead, in the generator's own decision order, with the same weights as our guard. BubbleFormer~\cite{sun2023bubbleformer}, which generates its own rooms from a given boundary, did not learn our room programs (validation room-count error 3.3--3.5 rooms, success 0.000) and is left out as a failed transfer.

\paragraph{Implementation.}
\Cref{tab:supp-hparams} lists the settings, fixed before evaluation; the status model uses the backbone's recipe unchanged. All runs used RTX PRO 6000 GPUs and CPU nodes on one cluster. Every result records its configuration, seeds, data and code fingerprints and checkpoint. Code will be made publicly available.

\begin{table}[!t]
  \centering\small
  \setlength{\tabcolsep}{3pt}
  \caption{\textbf{Settings} of our models. The plain and status models share them; the GRPO row applies only to the GRPO + guard baseline.}
  \label{tab:supp-hparams}
  \begin{tabular}{@{}lp{5.0cm}@{}}
    \toprule
    Item & Setting \\
    \midrule
    Encoder & Transformer over rooms, width 128, depth 2, 4 heads, FF 256, dropout 0.1, no positional encoding \\
    Training & batch 256, AdamW, lr $3\times10^{-4}$, wd $10^{-4}$, $\le$50 epochs, patience 8, clip 1.0; 3 seeds \\
    Briefs & 8 hindsight briefs per training plan, 0--8 atoms \\
    Sampling & $T{=}16$ steps, random order, temperature 1, 8 draws per program \\
    GRPO & 64 programs $\times$ group 8, $\beta=0.005$, 3{,}000 steps, lr $10^{-4}$ \\
    \bottomrule
  \end{tabular}
\end{table}

\section{All graph metrics and the ablations}
\label{sec:supp-results}

\paragraph{Protocol.}
Success is per draw: returned (not abstained, brief not flagged infeasible), valid and every atom met. Intervals resample programs with all their draws (plan bootstrap), so differences between methods on the same programs are paired. \emph{Bias} is the difference in mean access edges per graph, on hard briefs, between a certifier's returned graphs and exact samples of the same model's conditional, which are the model's own successful draws without a certifier. \emph{Effective sample size} is $(\sum_i w_i)^2/\sum_i w_i^2$ over a program's guarded draws, relative to their number. \emph{Realism} is MMD$^2$ with a Gaussian kernel ($\sigma=1$) on the total variation between per-graph histograms of access degree, wall clustering and the normalized-Laplacian spectrum of the wall graph. It compares the returned graphs with the real test graphs of the same plans, at most 1{,}500 graphs per side. We report it only where the brief has at least 5\% support, since rarer briefs ask for structure the real plan lacks. \emph{Cost} is single-core CPU seconds per graph on the same 24 programs.

\paragraph{Every metric in one table.}
\Cref{tab:supp-metrics} adds to \cref{tab:certifiers,tab:published} the metrics used for graph generators and link predictors. Per pair, F1 compares each graph with the program's real plan per relation, after the best relabeling of identical rooms. Per graph, \emph{distinct} counts different graphs (up to relabeling) among a program's 8 draws, \emph{novel} is the share of graphs absent from the complete training plans, and \emph{modal} the share equal to the program's most common training graph. AP scores a model's pair probabilities with nothing decided against the real test pairs. Three points stand out.
(i) Apart from the two baselines with reduced training budgets, generators without a certifier are valid in 0.79--0.93 of draws, close to the 0.96 of real graphs. What they miss is the brief, mostly the en-suite door (\cref{tab:supp-requests}): generators without a certifier meet B2 in 0.29--0.84 of draws and the status model in 0.85.
(ii) The status input does not buy success with diversity or memorization. The status model draws as many distinct graphs as the backbone (7.2 against 7.1 of 8), repeats the program's most common training graph less often (0.05 against 0.08; real test graphs 0.17), and its pair probabilities predict real edges as well (AP 0.748 against 0.746). Narrowing appears only for GRPO + guard and MILP 8-best (5.3 and 3.8 distinct graphs).
(iii) F1 measures closeness to one real layout, and that layout meets the brief in only half of the programs, so a method that meets the brief must often depart from it. Access F1 stays within 0.73--0.85 for every method. Passage F1 is lowest for repair-based pipelines (NSPSG 0.46, plain + repair 0.54, DiGress + repair 0.64) and higher for status + repair (0.72), whose draws need fewer edits.

\begin{table*}[!t]
  \centering\small
  \caption{\textbf{Every graph metric} (brief suite, three seeds). Valid, brief and success are per draw over all 1{,}988 programs, real test graphs included, with abstentions as failures. The other columns use the returned graphs of programs on which every compared method returns one (1{,}891; filters 1{,}883; Xia \etal 1{,}692). F1 is permutation-optimal against the program's real plan; Acc.\ edges is the mean number of access pairs. AP: the model's pair probabilities with nothing decided and each test plan's own hindsight brief, all 7{,}201 test plans, mean over corpus and relation; it is a property of the trained model, so it is given once per model. $^\dagger$Reduced training budget.}
  \label{tab:supp-metrics}
  \resizebox{\textwidth}{!}{%
\begin{tabular}{@{}lcccccccccccc@{}}
\toprule
 & \multicolumn{3}{c}{Per draw, all programs} & \multicolumn{3}{c}{Diversity and memorization} & \multicolumn{5}{c}{Per pair: F1 to the real plan, edges} & Link pred. \\
\cmidrule(lr){2-4}\cmidrule(lr){5-7}\cmidrule(lr){8-12}\cmidrule(l){13-13}
Method & Valid & Brief & Success & Distinct & Novel & Modal & Wall & Door & Passage & Access & Acc.\ edges & AP \\
\midrule
Real test graphs & 0.96 & 0.47 & 0.456 & -- & 0.39 & 0.17 & 1 & 1 & 1 & 1 & 5.2 & -- \\
\midrule
\multicolumn{13}{@{}l}{\emph{Without a certifier}} \\
GraphRNN & 0.90 & 0.42 & 0.371 & 6.6 & 0.52 & 0.15 & 0.79 & 0.83 & 0.84 & 0.85 & 5.2 & -- \\
DGMG$^\dagger$ & 0.51 & 0.19 & 0.125 & 7.3 & 0.79 & 0.06 & 0.75 & 0.72 & 0.73 & 0.73 & 5.1 & -- \\
Xia \etal$^\dagger$ & 0.64 & 0.28 & 0.192 & 7.3 & 0.74 & 0.07 & 0.72 & 0.75 & 0.77 & 0.77 & 5.0 & -- \\
DiGress & 0.92 & 0.73 & 0.675 & 6.9 & 0.58 & 0.07 & 0.78 & 0.76 & 0.84 & 0.79 & 5.2 & -- \\
ConStruct & 0.93 & 0.56 & 0.512 & 6.6 & 0.51 & 0.14 & 0.79 & 0.82 & 0.83 & 0.84 & 5.2 & -- \\
Plain (ours, backbone) & 0.79 & 0.51 & 0.422 & 7.1 & 0.68 & 0.08 & 0.77 & 0.79 & 0.81 & 0.80 & 5.2 & 0.746 \\
\textbf{Status (ours)} & 0.83 & 0.76 & 0.675 & 7.2 & 0.73 & 0.05 & 0.77 & 0.75 & 0.82 & 0.77 & 5.3 & 0.748 \\
\midrule
\multicolumn{13}{@{}l}{\emph{With an exact certifier}} \\
GraphRNN + guard & 0.94 & 0.94 & 0.938 & 6.6 & 0.74 & 0.04 & 0.78 & 0.77 & 0.79 & 0.80 & 5.7 & -- \\
DiGress + repair & 1.00 & 0.95 & 0.951 & 6.9 & 0.66 & 0.04 & 0.78 & 0.75 & 0.64 & 0.79 & 5.6 & -- \\
NSPSG & 1.00 & 0.95 & 0.951 & 6.9 & 0.80 & 0.04 & 0.79 & 0.78 & 0.46 & 0.81 & 5.8 & -- \\
MILP 8-best & 0.94 & 0.94 & 0.944 & 3.8 & 0.86 & 0.01 & 0.79 & 0.76 & 0.82 & 0.79 & 6.1 & -- \\
Retrieve-and-filter & 0.81 & 0.81 & 0.808 & -- & -- & -- & -- & -- & -- & -- & -- & -- \\
GRPO + guard & 0.95 & 0.95 & 0.949 & 5.3 & 0.75 & 0.09 & 0.81 & 0.79 & 0.87 & 0.82 & 6.0 & 0.718 \\
Plain + filter & 0.92 & 0.92 & 0.921 & 7.0 & 0.74 & 0.03 & 0.76 & 0.72 & 0.82 & 0.76 & 5.4 & -- \\
Plain + repair & 1.00 & 0.95 & 0.950 & 7.1 & 0.85 & 0.03 & 0.78 & 0.75 & 0.54 & 0.79 & 5.7 & -- \\
Plain + guard & 0.92 & 0.92 & 0.925 & 6.9 & 0.79 & 0.03 & 0.77 & 0.74 & 0.82 & 0.78 & 5.6 & -- \\
\textbf{Status + filter} & 0.95 & 0.95 & 0.948 & 7.0 & 0.75 & 0.03 & 0.77 & 0.74 & 0.83 & 0.77 & 5.5 & -- \\
\textbf{Status + repair} & 0.99 & 0.95 & 0.950 & 7.2 & 0.78 & 0.03 & 0.77 & 0.74 & 0.72 & 0.77 & 5.6 & -- \\
\textbf{Status + guard} & 0.92 & 0.92 & 0.921 & 6.8 & 0.76 & 0.03 & 0.77 & 0.74 & 0.83 & 0.77 & 5.5 & -- \\
\bottomrule
\end{tabular} }
\end{table*}

\begin{table*}[!t]
  \centering\small
  \caption{\textbf{Requests met without a certifier} (share of draws meeting each request where it applies, both corpora, three seeds; real test graphs on the same programs).}
  \label{tab:supp-requests}
  \resizebox{0.6\textwidth}{!}{%
\begin{tabular}{@{}lcccccc@{}}
\toprule
Graphs & B1 & B2 & B3 & B4 & B5 & B6 \\
\midrule
Real test graphs & 0.93 & 0.49 & 0.99 & 0.78 & 0.99 & 0.94 \\
GraphRNN & 0.94 & 0.45 & 0.97 & 0.95 & 0.78 & 0.96 \\
DGMG$^\dagger$ & 0.84 & 0.29 & 0.85 & 0.89 & 0.74 & 0.70 \\
Xia \etal$^\dagger$ & 0.84 & 0.41 & 0.89 & 0.93 & 0.72 & 0.78 \\
DiGress & 0.99 & 0.84 & 0.99 & 1.00 & 0.79 & 1.00 \\
ConStruct & 1.00 & 0.59 & 0.98 & 1.00 & 0.77 & 0.99 \\
Plain (ours, backbone) & 0.97 & 0.63 & 0.98 & 0.98 & 0.80 & 0.85 \\
\textbf{Status (ours)} & 0.98 & 0.85 & 0.99 & 0.99 & 0.87 & 0.91 \\
\bottomrule
\end{tabular} }
\end{table*}

\paragraph{Which part of the status input?}
The status input has two parts: the \textsc{met} and \textsc{dead} flags of the requirements, and the allowed outcomes of each undecided pair. We retrained the status model with one part hidden, in training and at inference, with the backbone's recipe and seeds. \emph{Flags only} sets every allowed bit to zero. \emph{Allowed only} sets the flags to zero but keeps the requirement tokens, so the model still sees which requirements are in force. No requirement is \textsc{dead} in training (\cref{lem:hindsight}), so the model learns from the \textsc{met} flags only. A dead requirement also shows in the allowed bits, since it leaves no outcome allowed. Retraining the full status model with the same code reproduced its checkpoints bit for bit, so the four models of \cref{tab:supp-parts} come from one code state. The allowed outcomes carry most of the gain: on hard briefs sampling succeeds in 0.409 of draws without the flags and in 0.125 without the allowed outcomes, against 0.467 with both and 0.105 for the plain model. The flags add little alone ($+0.020$ [0.015, 0.025] over plain) but $+0.058$ [0.051, 0.064] on top of the allowed outcomes, positive in every seed. Under repair, edits fall from 0.70 per graph (plain) to 0.63, 0.44 and 0.38.

\begin{table*}[!t]
  \centering\small
  \caption{\textbf{Status-input ablation.} Sampling success on hard briefs (95\% interval, seed range), all programs and the mined suite; exact repair: edits per graph, share of draws changed, in-support access MMD$\times10^3$.}
  \label{tab:supp-parts}
  \resizebox{0.75\textwidth}{!}{%
\begin{tabular}{@{}ccccccccc@{}}
\toprule
\multicolumn{2}{c}{Status input} & \multicolumn{4}{c}{Sampling success} & \multicolumn{3}{c}{Exact repair} \\
\cmidrule(r){1-2}\cmidrule(lr){3-6}\cmidrule(l){7-9}
Flags & Allowed & Hard & Seeds & All & Mined & Edits & Changed & Acc.\ MMD \\
\midrule
-- & -- & 0.105\,{\scriptsize[0.095, 0.115]} & 0.10--0.11 & 0.422 & 0.700 & 0.70 & 0.53 & 18.0 \\
\checkmark & -- & 0.125\,{\scriptsize[0.114, 0.136]} & 0.12--0.14 & 0.464 & 0.726 & 0.63 & 0.49 & 15.6 \\
-- & \checkmark & 0.409\,{\scriptsize[0.389, 0.431]} & 0.33--0.49 & 0.633 & 0.784 & 0.44 & 0.32 & 12.5 \\
\checkmark & \checkmark & 0.467\,{\scriptsize[0.444, 0.491]} & 0.38--0.56 & 0.675 & 0.792 & 0.38 & 0.27 & 11.0 \\
\bottomrule
\end{tabular} }
\end{table*}

\paragraph{Where the status helps.}
\Cref{tab:strata} breaks success down by support. Without a certifier the status input helps most where few training plans meet the brief, and under filtering where none does.

\begin{table*}[!t]
  \centering\small
  \caption{\textbf{Success by support of the brief} (both corpora, with the number of programs and of feasible briefs). The hard cell is the MagicPlan part of the first two columns.}
  \label{tab:strata}
  \resizebox{0.6\textwidth}{!}{%
\begin{tabular}{@{}lcccc@{}}
\toprule
Support of the brief & none & $<$5\% & 5--50\% & $\ge$50\% \\
Plans (feasible) & 414 (317) & 180 (180) & 496 (496) & 898 (898) \\
\midrule
Plain, sampling & 0.097 & 0.129 & 0.378 & 0.655 \\
Status, sampling & 0.403 & 0.614 & 0.765 & 0.764 \\
Plain, filter & 0.639 & 0.957 & 1.000 & 1.000 \\
Status, filter & 0.749 & 1.000 & 1.000 & 1.000 \\
Plain, guard & 0.730 & 0.965 & 0.982 & 0.975 \\
Status, guard & 0.730 & 0.958 & 0.980 & 0.968 \\
\bottomrule
\end{tabular} }
\end{table*}

\paragraph{An expressive-encoder control.}
\Cref{prop:invisible} does not cover encoders that update pair states, so we test one. We add four PPGN blocks~\cite{maron2019provably} to the backbone. After the room Transformer, they update a tensor of room pairs by multiplying pair matrices over a third room (the mechanism of 2-FWL), and their output replaces the pair-state embedding in the pair head (1.04M parameters against 0.51M). Unlike the Transformer's, its pair scores can separate the bridge of $s_B$ from the other undecided pairs. Trained with the backbone's recipe, it fits real plans slightly better (validation AP 0.744 against 0.738) but does not meet hard briefs more often (0.112 against 0.105), while the status input lifts it to 0.503 and acts under every certifier as it does on the Transformer (\cref{tab:supp-encoder}).

\begin{table*}[!t]
\centering\small
\caption{\textbf{Expressive-encoder control} (hard briefs, three seeds). Draws: generate-and-filter draws per sample on feasible hard briefs (an abstention counts its 64). Edits: pairs changed by repair per graph. Acc.\ MMD: access-degree MMD$^2\times10^3$ of repaired graphs on in-support briefs. Bias: extra access edges per guarded graph against the same model's exact samples.}
\label{tab:supp-encoder}
\resizebox{\textwidth}{!}{%
\begin{tabular}{@{}llcccccccc@{}}
\toprule
 & & & Sampling & \multicolumn{2}{c}{Filter ($\le$64)} & \multicolumn{2}{c}{Exact repair} & \multicolumn{2}{c}{Status guard} \\
\cmidrule(lr){4-4}\cmidrule(lr){5-6}\cmidrule(lr){7-8}\cmidrule(l){9-10}
Encoder & Model & Params & Success & Success & Draws & Edits & Acc.\ MMD & Success & Bias \\
\midrule
Transformer & plain & 0.51M & 0.105 & 0.732 & 17.6 & 0.70 & 18.0 & 0.799 & +0.52 \\
Transformer & status & 0.51M & 0.467 & 0.822 & 3.3 & 0.38 & 11.0 & 0.798 & +0.13 \\
PPGN pairs & plain & 1.04M & 0.112 & 0.730 & 17.5 & 0.55 & 13.6 & 0.810 & +0.48 \\
PPGN pairs & status & 1.04M & 0.503 & 0.811 & 4.2 & 0.29 & 9.2 & 0.805 & +0.14 \\
\bottomrule
\end{tabular} }
\end{table*}

\paragraph{The held-out brief.}
The held-out brief (\cref{tab:supp-briefs}) is evaluated with the same checkpoints, certifiers and seeds (\cref{tab:supp-heldout}). On MagicPlan the status input again lifts sampling where the brief is rare (no support: 0.081 to 0.205; under 5\%: 0.127 to 0.357). Generate-and-filter certifies within 64 draws 0.37 to 0.72 of draws where no training plan meets the brief (\cref{sec:exp-ablation}), and 0.41 to 0.74 on all MagicPlan programs with support below 5\% (\cref{tab:supp-heldout} pools both corpora). No ResPlan program has a training plan that meets this brief, because ResPlan's living rooms are hubs, which goes against H4 and H5. On ResPlan, sampling stays near zero with either model, and only the certifiers meet the brief: exact repair certifies 93\% of draws, while the guard abstains on 41--44\%, because these requests combine into dead ends that per-requirement status does not reveal (\cref{sec:conclusion}).

\begin{table*}[!t]
\centering\small
\caption{\textbf{Held-out brief} (H1--H6; three seeds). Success on the 1{,}976 programs it applies to and on its hard cell (support below 5\%, both corpora, 1{,}093 programs; 95\% interval); abstention; filter draws per sample; repair edits per graph.}
\label{tab:supp-heldout}
\resizebox{0.75\textwidth}{!}{%
\begin{tabular}{@{}llccccc@{}}
\toprule
Certifier & Model & Success (all) & Success (hard) & Abst. & Draws & Edits \\
\midrule
Sampling only & plain & 0.334 & 0.021\,{\scriptsize[0.016, 0.026]} & 0.000 & -- & -- \\
 & status & 0.405 & 0.055\,{\scriptsize[0.044, 0.066]} & 0.000 & -- & -- \\
Generate-and-filter ($\le$64) & plain & 0.503 & 0.102\,{\scriptsize[0.085, 0.120]} & 0.497 & 33.1 & -- \\
 & status & 0.587 & 0.254\,{\scriptsize[0.232, 0.275]} & 0.413 & 29.2 & -- \\
Exact min-edit repair & plain & 0.970 & 0.945\,{\scriptsize[0.935, 0.955]} & 0.000 & -- & 2.26 \\
 & status & 0.969 & 0.943\,{\scriptsize[0.932, 0.953]} & 0.000 & -- & 1.85 \\
Status guard & plain & 0.798 & 0.647\,{\scriptsize[0.626, 0.667]} & 0.202 & -- & -- \\
 & status & 0.810 & 0.670\,{\scriptsize[0.650, 0.689]} & 0.190 & -- & -- \\
Retrieve-and-filter & -- & 0.456 & 0.021\,{\scriptsize[0.013, 0.029]} & 0.544 & -- & -- \\
\bottomrule
\end{tabular} }
\end{table*}

\section{Generated room graphs}
\label{sec:supp-qual}

\begin{figure*}[!t]
  \centering
  \includegraphics[width=\textwidth]{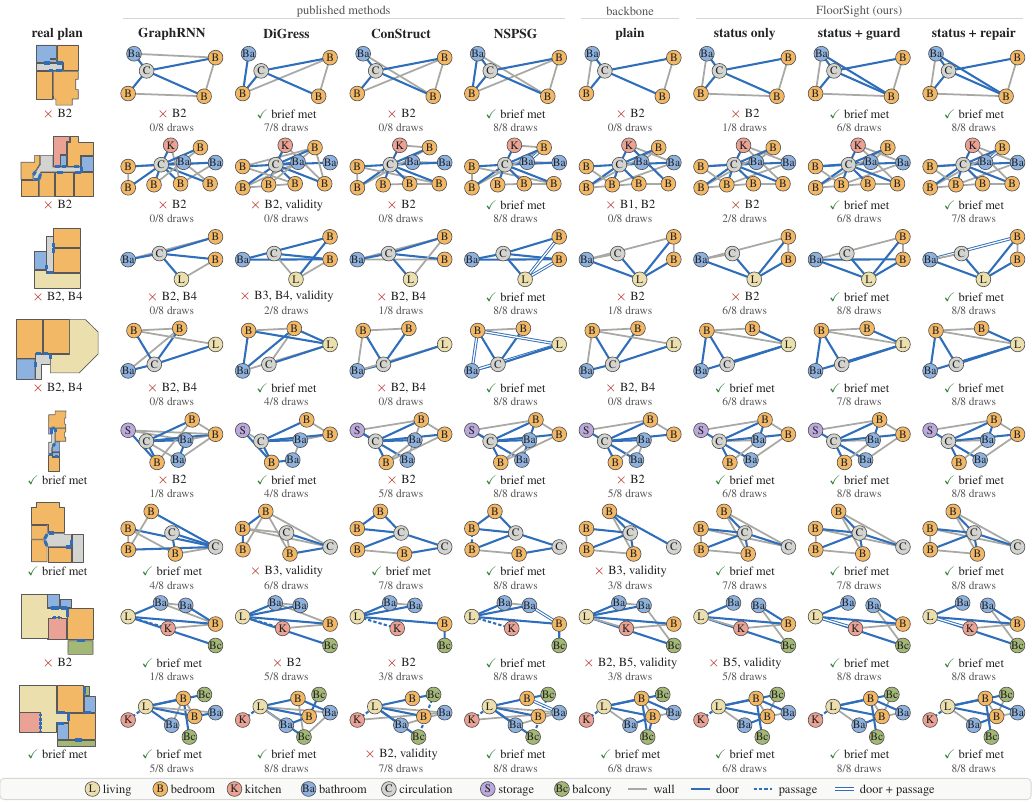}
  \caption{\textbf{Room graphs for random programs} (generator seed 101), from programs with no support for the brief (top) to programs whose brief most training plans meet (bottom; the last two rows are ResPlan). Each method's first returned graph, the requests it violates and how many of its 8 draws meet brief and validity.}
  \label{fig:supp-gallery}
\end{figure*}

\paragraph{Examples of the main paper.}
\Cref{fig:qual-graphs,fig:qual-drawings} show three programs selected from the hard cell (MagicPlan, support below 5\%, feasible brief, 5--10 rooms; 367 programs). Candidates needed a clean real plan (at least 95\% axis-aligned walls, no room thinner than 11\% of the plan or under 2\% of its area, aspect 0.55--1.8, at most 10 corners per room, rooms filling at least 75\% of the bounding box) and outcomes in line with the reported rates. These examples therefore favor our method by construction.

\paragraph{Random programs.}
\Cref{fig:supp-gallery} instead shows programs drawn by a fixed seed without looking at any generated graph: two from each MagicPlan stratum below 5\% and one from each other stratum with enough programs, among programs with a feasible brief and 5--10 rooms. Each cell is the method's first returned graph. Published generators often return plausible graphs that miss the en-suite door (B2). Without a certifier, the status model meets the brief at least as often as the backbone for every program, and the certified methods meet it in every returned graph; status + guard succeeds in fewer than 8 of 8 draws only where it abstains, which counts as a failure.

\section{Downstream floorplans}
\label{sec:supp-downstream}

\begin{table*}[!t]
  \centering\small
  \caption{\textbf{HouseDiffusion drawing from each graph source} (1{,}891 programs with a feasible brief; 4 drawings per graph, 3 generator seeds; HD training seed 1). Brief: every atom of the brief on the input graph and on the graph read back from the drawn plan. Valid: all validity requirements after drawing.}
  \label{tab:supp-downstream}
  \resizebox{\textwidth}{!}{%
\begin{tabular}{@{}lcccccccc@{}}
\toprule
Graph source & Brief (graph) & Valid (drawn) & Brief (drawn) & Compat.\ F1 & Overlap $\downarrow$ & All rooms & KID$\times10^3$ & FID \\
\midrule
Real graph & 0.494 & 0.238 & 0.225 & 0.836 & 0.337 & 0.824 & 41.8 & 38.3 \\
Plain, sampling & 0.539 & 0.206 & 0.257 & 0.812 & 0.439 & 0.831 & 44.9 & 40.6 \\
Plain + guard & 1.000 & 0.244 & 0.481 & 0.812 & 0.455 & 0.830 & 46.8 & 41.9 \\
Plain + exact repair & 1.000 & 0.253 & 0.481 & 0.811 & 0.455 & 0.829 & 47.4 & 42.2 \\
NSPSG & 1.000 & 0.261 & 0.488 & 0.820 & 0.428 & 0.829 & 45.4 & 40.9 \\
DiGress & 0.772 & 0.227 & 0.405 & 0.829 & 0.382 & 0.828 & 42.1 & 38.6 \\
MILP 8-best & 1.000 & 0.251 & 0.482 & 0.790 & 0.520 & 0.829 & 58.5 & 50.1 \\
\textbf{Status + guard} & 1.000 & 0.237 & 0.485 & 0.811 & 0.451 & 0.828 & 46.3 & 41.5 \\
\textbf{Status + exact repair} & 0.999 & 0.238 & 0.486 & 0.810 & 0.453 & 0.827 & 46.3 & 41.6 \\
\bottomrule
\end{tabular} }
\end{table*}

\begin{figure*}[!t]
  \centering
  \includegraphics[width=0.86\textwidth]{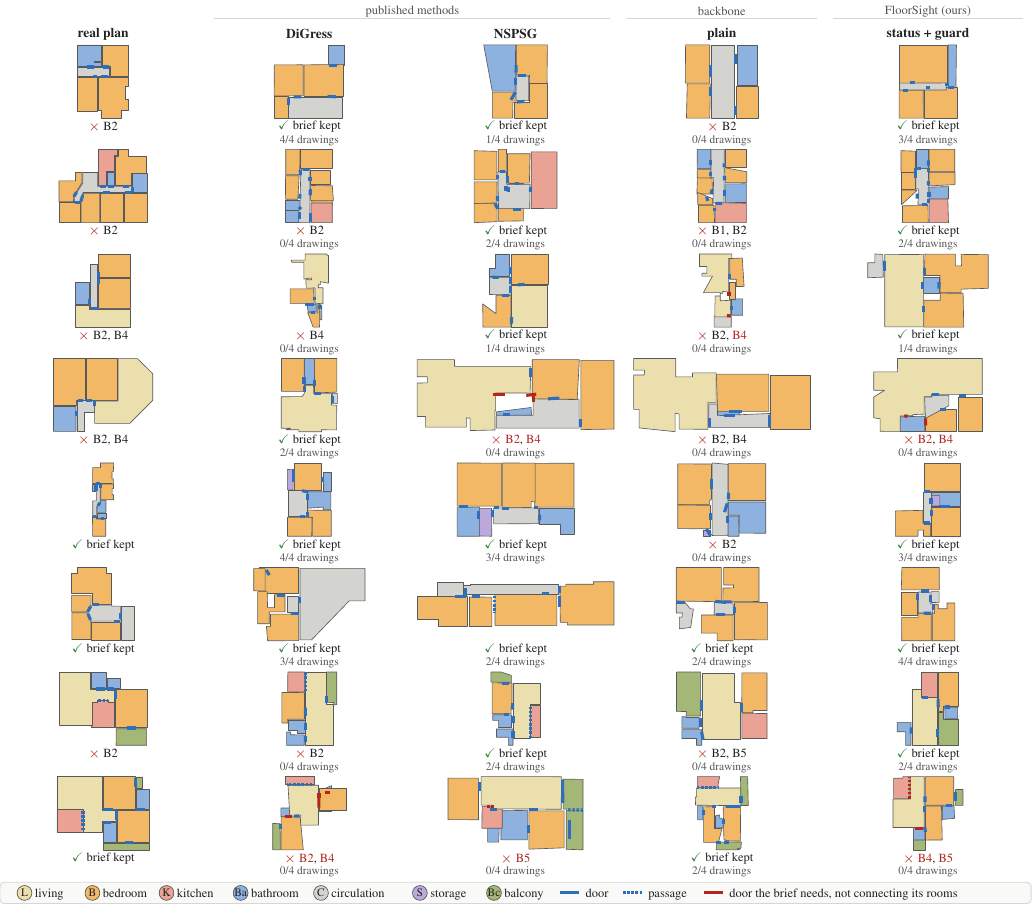}
  \caption{\textbf{HouseDiffusion drawings of the graphs of \cref{fig:supp-gallery}.} Each graph is drawn four times, and each cell shows the first drawing keeping the brief, or else the first. Footers give the requests violated, black if the graph misses them and red if drawing loses them, and how many of the four drawings keep the brief. Red bars are doors a lost request needs that the drawing does not connect to both rooms.}
  \label{fig:supp-drawings}
\end{figure*}

\paragraph{HouseDiffusion on our corpus.}
We retrain HouseDiffusion~\cite{shabani2023housediffusion} on our training split and change only its graph interface. A plan is a set of elements: its rooms, plus one opening element per access pair, typed door, passage or \emph{unspecified}. The graph enters only through attention masks: the corner tokens of two rooms attend to each other iff the rooms share a wall, and an opening's tokens attend to its two rooms. Relative to the official code, we use our 11 room classes plus the three opening types and no front door, and these adjacency and access masks instead of HouseDiffusion's single door mask. Each training seed runs 250k steps (batch 512).

\paragraph{Drawing and reading back.}
For the 1{,}891 programs with a feasible brief, the first returned graph of each method and generator seed is drawn $K{=}4$ times. The number of corners per room is sampled from training histograms with a random generator keyed by plan and drawing, so every graph source gets the same counts for the same program. Drawn plans are read back with the interface rule of \cref{eq:interface} and a single gap of 0.1 normalized units (interface $\ge0.04$), calibrated on validation geometry; on real test geometry this round trip reproduces 99.7\% of recorded adjacency pairs in both sources and 99.9--100\% of recorded openings, which bounds extraction error for every source.

\paragraph{Metrics.}
A drawing keeps the brief when its read-back graph meets it. We also report validity after drawing, compatibility F1 between input and read-back graph, any material room overlap, every requested room present, and KID~\cite{binkowski2018demystifying} and FID~\cite{heusel2017gans}. For these, every plan, real or drawn, is rendered by one renderer at $256\times256$ (rooms filled with their class color, outlined, openings as white segments) and embedded with the standard InceptionV3 pool features (2048-d); KID is the unbiased estimate with a cubic polynomial kernel over 100 subsets of 1{,}000, against the 2{,}074 door/passage-complete real test plans. \Cref{tab:supp-downstream} gives every graph source.

\paragraph{How drawing loses a requirement.}
An opening of the input graph is \emph{critical} for the brief when removing it alone (keeping the wall) makes an atom that the input meets fail. In the separate set of $K{=}16$ drawings per graph (generator seed 101, fresh diffusion noise), we take the drawings that lose a brief their certified input met although every room is drawn. In 89--92\% of them (by graph source) a critical opening is not kept. Drawing keeps critical openings barely more often than the others, since HD never sees the brief. A second HD training seed gives the same picture: certified sources keep the brief in 0.49--0.51 of drawings against 0.25 for plain sampling and for the real graphs. On the mined suite, where every brief is the hindsight brief of the program's own real plan, even the real graphs keep their own brief in only 0.52 of drawings, so geometry loses about half of a brief whatever its source.

\paragraph{Screening drawings.}
Since the brief is checked exactly on the read-back graph, a user can draw a certified graph several times and keep a drawing that passes. Checking validity as well, the first 16 drawings of a certified graph contain a plan that meets the brief and every validity requirement for 0.73--0.78 of programs, against 0.34 for real graphs.

\paragraph{Drawn plans for random programs.}
\Cref{fig:supp-drawings} draws the graphs of \cref{fig:supp-gallery}. Certified graphs give HD the openings the brief needs, but HD may still drop them (red).

\end{document}